%% file: main.tex
\documentclass{article}

\usepackage[accepted]{aistats2020}

\usepackage[round]{natbib}

\usepackage[utf8]{inputenc} %
\usepackage[T1]{fontenc}    %
\usepackage{hyperref}       %
\usepackage{url}            %
\usepackage{booktabs}       %
\usepackage{amsfonts}       %
\usepackage{nicefrac}       %
\usepackage{microtype}      %
\usepackage{bm}
\usepackage[page]{appendix}

\usepackage{definitions}
\hypersetup{
    colorlinks,
    linkcolor={red!90!black},
    citecolor={blue!50!black},
    urlcolor={blue!80!black}
}

\newcommand{\pt}{p_\thetab}
\newcommand{\qp}{q_\phib}

\renewcommand{\NN}{\mathcal{N}} %
\renewcommand{\ZZ}{\mathcal{Z}} %
\newcommand{\pftl}{p_{\fb, \Tb, \lambdab}}

\newcommand{\VAEICA}{iVAE }

\begin{document}

\twocolumn[

\aistatstitle{Variational Autoencoders and Nonlinear ICA:\\A Unifying Framework}

\aistatsauthor{ Ilyes Khemakhem \And Diederik P. Kingma \And Ricardo Pio Monti \And  Aapo Hyv{\"a}rinen }

\aistatsaddress{ Gatsby Unit\\ UCL \And  Google Brain \And Gatsby Unit \\ UCL \And Universit\'e Paris-Saclay,\\Inria, Univ.\ of Helsinki } 
]

\begin{abstract}
The framework of variational autoencoders allows us to efficiently learn deep latent-variable models, such that the model's marginal distribution over observed variables fits the data. Often, we're interested in going a step further, and want to approximate the true joint distribution over observed and latent variables, including the true prior and posterior distributions over latent variables. This is known to be generally impossible due to unidentifiability of the model. We address this issue by showing that for a broad family of deep latent-variable models, identification of the true joint distribution over observed and latent variables is actually possible up to  very simple transformations, thus achieving a principled and powerful form of disentanglement. Our result requires a factorized prior distribution over the latent variables that is conditioned on an additionally observed variable, such as a class label or almost any other observation. We build on recent developments in nonlinear ICA, which we extend to the case with noisy or undercomplete observations, integrated in a maximum likelihood framework. The result also trivially contains identifiable flow-based generative models as a special case.

\end{abstract}

\input{sections/1_introduction.tex}
\input{sections/2_unidentifiability.tex}

\input{sections/3_model.tex}

\input{sections/4_technical_identifiability.tex}

\input{sections/5_experiments.tex}
\input{sections/6_conclusion.tex}

\bibliographystyle{apalike}
\bibliography{refs}

\input{sections/7_appendix_proofs.tex}

\input{sections/8_appendix_bis.tex}
\input{sections/9_appendix_figs.tex}

\end{document}

%% file: sections/1_introduction.tex
\section{INTRODUCTION}
\label{introduction}

The framework of variational autoencoders \citep{kingma2013autoencoding,rezende2014stochastic} (VAEs) and its extensions (e.g. \cite{burda2015importance,kingma2016improving,tucker2018doubly,maaloe2019biva}) offers a scalable set of techniques for learning deep latent-variable models and corresponding inference models.
With VAEs, we can in principle learn flexible models of data such that, after optimization, the model's implicit marginal distribution over the observed variables approximates their true (but unknown) distribution. With VAEs we can also efficiently synthesize pseudo-data from the model.

However, we're often interested in going a step further and want to learn the true joint distribution over both observed and latent variables. This is generally a very difficult task, since by definition we only ever observe the observed variables, never the latent variables, therefore we cannot directly estimate their joint distribution. If we could however somehow achieve this task and learn the true joint distribution, this would imply that we have also learned to approximate the true prior and posterior distributions over latent variables. Learning about these distributions can be very interesting for various purposes, for example in order to learn about latent structure behind the data, or in order to infer the latent variables from which the data originated.

Learning the true joint distribution is only possible when the model is \emph{identifiable}, as we will explain. The original VAE theory doesn't tell us when this is the case; it only tells us how to optimize the model's parameters such that its (marginal) distribution over the observed variables matches the data. The original theory doesn't tell us if or when we learn the correct joint distribution over observed and latent variables.

Almost no literature exists on achieving this goal. A pocket of the VAE literature works towards the related goal of \emph{disentanglement}, but offers no proofs or theoretic guarantees of identifiability of the model or its latent variables. The most prominent of such models are $\beta$-VAEs and their extensions \citep{burgess2018understanding,higgins2016betavae,higgins2018definition,esmaeili2018structured,kim2018disentangling,chen2018isolating}, in which the authors introduce adjustable hyperparameters in the VAE objective to encourage disentanglement. Other work attempts to find maximally independent components through the GAN framework \citep{brakel2017learning}. However, models in these earlier works are actually non-identifiable due to non-conditional latent priors, as has been seen empirically \citep{locatello2018challenging}, and we will show formally below.

Recent work in nonlinear Independent Component Analysis (ICA) theory \citep{hyvarinen2016unsupervised,hyvarinen2017nonlinear,hyvarinen2019nonlinear} provided the first identifiability results for deep latent-variable models. Nonlinear ICA provides a rigorous framework for recovering independent latents that were transformed by some invertible nonlinear transformation into the data.  Some special but not very restrictive conditions are necessary, since it is known that when the function from latent to observed variables is nonlinear, the general problem is ill-posed, and one cannot recover the independent latents \citep{hyvarinen1999nonlinear}. However, existing nonlinear ICA methods do not learn to model the data distribution (pdf), nor do they allow us to synthesize pseudo-data.

In this paper we show that under relatively mild conditions the joint distribution over observed and latent variables in VAEs is identifiable and learnable, thus bridging the gap between VAEs and nonlinear ICA.
To this end, we establish a principled connection between VAEs and an identifiable nonlinear ICA model, providing a unified view of two complementary methods in unsupervised representation learning. This integration is achieved by using a latent prior that has a factorized distribution that is conditioned on additionally observed variables, such as a class label, time index, or almost any other further observation. Our theoretical results trivially apply to any consistent parameter estimation method for deep latent-variable models, not just the VAE framework. We found the VAE a logical choice since it allows for efficient latent-variable inference and scales to large datasets and models.
Finally, we put our theoretical results to a test in experiments. Perhaps most notably, we find that on a synthetic dataset with known ground-truth model, our method with an identifiable VAE indeed learns to closely approximate the true joint distribution over observed and latent variables, in contrast with a baseline non-identifiable model.

%% file: sections/2_unidentifiability.tex
\section{UNIDENTIFIABILITY OF DEEP LATENT VARIABLE MODELS}
\label{unidentifiability}

\subsection{Deep latent variable models}

Consider an observed data variable (random vector) $\xb \in  \RR^d$, and a latent random vector $\zb \in \RR^n$. A common deep latent variable model has the following structure:
\begin{equation}
\label{eq:generalgen}
    p_{\thetab}(\xb, \zb) = p_{\thetab}(\xb\vert\zb)p_{\thetab}(\zb)
\end{equation}
where $\thetab \in \Theta$ is a vector of parameters, $\pt(\zb)$ is called a prior distribution over the latent variables. The distribution $\pt(\xb|\zb)$, often parameterized with a neural network called the \emph{decoder}, tells us how the distribution on $\xb$ depends on the values of $\zb$. The model then gives rise to the observed distribution of the data as:
\begin{equation}
p_{\thetab}(\xb)=\int p_{\thetab}(\xb, \zb) \dd\zb
\label{eq:marginal}
\end{equation}
Assuming $p_{\thetab}(\xb|\zb)$ is modelled by a deep neural network, this can model a rich class of data distributions $p_\thetab(\xb)$. 

We assume that we observe data which is generated from an underlying joint distribution $p_{\thetab^*}(\xb, \zb) = p_{\thetab^*}(\xb\vert\zb)p_{\thetab^*}(\zb) $ where $\thetab^*$ are its true but unknown parameters. We then collect a dataset of observations of $\xb$:
\begin{align*}
    \mathcal{D} = \{\xb^{(1)}, \ldots, \xb^{(N)}\} \text{\;\;where\;\;} & \zb^{*(i)} \sim p_{\thetab^*}(\zb)\\ & \xb^{(i)} \sim p_{\thetab^*}(\xb\vert\zb^{*(i)})
\end{align*}
Note that the original values $\zb^{*(i)}$ of the latent variables $\zb$ are by definition not observed and unknown. The ICA literature, including this work, uses the term \emph{sources} to refer to $\zb^{*(i)}$.
Also note that we could just as well have written: $\xb^{(i)} \sim p_{\thetab^*}(\xb)$.

The VAE framework~\citep{kingma2013autoencoding,rezende2014stochastic} allows us to efficiently optimize the parameters $\thetab$ of such models towards the (approximate) maximum marginal likelihood objective, such that after optimization:
\begin{align}
    p_{\thetab}(\xb) \approx p_{\thetab^*}(\xb)
    \label{eq:vae_approx}
\end{align}
In other words, after optimization we have then estimated the marginal density of $\xb$.

\subsection{Parameter Space vs Function Space}
\label{sec:functionspace}

In this work we use slightly non-standard notation and nomenclature: we use $\thetab \in \Theta$ to refer to the model parameters in \emph{function space}. In contrast, let $\mathbf{w} \in W$ refer to the space of original neural network parameters (weights, biases, etc.) in which we usually perform gradient ascent.  

\subsection{Identifiability}
\label{sec:identifiability}
The VAE model actually learns a full generative model $p_{\thetab}(\xb, \zb) = p_{\thetab}(\xb\vert\zb)p_{\thetab}(\zb)$ and an inference model $q_\phib(\zb|\xb)$ that approximates its posterior $p_\thetab(\zb|\xb)$. The problem is that we generally have no guarantees about what these learned distributions actually are: all we know is that the marginal distribution over $\xb$ is meaningful (Eq.~\ref{eq:vae_approx}). The rest of the learned distributions are, generally, quite meaningless.

What we are looking for is models for which the following implication holds for all $(\xb, \zb)$:
\begin{align}
\forall (\thetab,\thetab'):\;\; p_{\thetab}(\xb) = p_{\thetab'}(\xb) \;\;\implies\;\; \thetab = \thetab'
\label{eq:identifiable}
\end{align}
That is: if any two different choices of model parameter $\thetab$ and $\thetab'$ lead to the same marginal density $p_\thetab(\xb)$, then this would imply that they are equal and thus have matching joint distributions $p_\thetab(\xb,\zb)$. This means that if we learn a parameter $\thetab$ that fits the data perfectly: $p_{\thetab}(\xb) = p_{\thetab^*}(\xb)$ (the ideal case of Eq.~\ref{eq:vae_approx}), then its joint density also matches perfectly: $p_{\thetab}(\xb, \zb) = p_{\thetab^*}(\xb, \zb)$. If the joint density matches, this also means that we found the correct prior $\pt(\zb) = p_{\thetab^*}(\zb)$ and correct posteriors $\pt(\zb\vert\xb) = p_{\thetab^*}(\zb\vert\xb)$. In case of VAEs, we can then also use the inference model $\qp(\zb|\xb)$ to efficiently perform inference over the sources $\zb^*$ from which the data originates.

The general problem here is a lack of \emph{identifiability} guarantees of the deep latent-variable model. We illustrate this by showing that any model with unconditional latent distribution $\pt(\zb)$ is unidentifiable, i.e. that Eq.~(\ref{eq:identifiable}) does not hold. In this case, we can always find transformations of $\zb$ that changes its value but does not change its distribution. For a spherical Gaussian distribution $\pt(\zb)$, for example, applying a rotation keeps its distribution the same. We can then incorporate this transformation as the first operation in $\pt(\xb\vert\zb)$. This will not change $\pt(\xb)$, but it will change $\pt(\zb\vert\xb)$, since now the values of $\xb$ come from different values of $\zb$.  %
This is an example of a broad class of commonly used models that are non-identifiable. We show rigorously in Supplementary Material~\ref{app:unident} that, in fact, models with \textit{any} form of unconditional prior $\pt(\zb)$ are unidentifiable.

%% file: sections/3_model.tex
\section{AN IDENTIFIABLE MODEL BASED ON CONDITIONALLY FACTORIAL PRIORS}

In this section, we define a broad family of deep latent-variable models which is identifiable, and we show how to estimate the model and its posterior through the VAE framework. We call this family of models, together with its estimation method, Identifiable VAE, or \VAEICA for short.

\subsection{Definition of proposed model}
\label{sec:model}

The primary assumption leading to identifiability is a conditionally factorized prior distribution over the latent variables $\pt(\zb|\ub)$, where $\ub$ is an additionally observed variable \citep{hyvarinen2019nonlinear}. The variable $\ub$ could be, for example, the time index in a time series \citep{hyvarinen2016unsupervised}, previous data points in a time series, some kind of (possibly noisy) class label, or another concurrently observed variable.

Formally, let $\xb \in  \RR^d$,  and $\ub \in  \RR^m$ be two observed random variables, and $\zb \in  \RR^n$ (lower-dimensional, $n \leq d$) a latent variable. Let $\thetab = (\fb, \Tb, \lambdab)$ be the parameters of the following conditional generative model:
\begin{equation}
\label{eq:gen}
    \pt(\xb, \zb \vert \ub) = p_\fb(\xb\vert\zb)p_{\Tb, \lambdab}(\zb\vert\ub)
\end{equation}
where we first define:
\begin{equation}    
\label{eq:xcz}
    p_\fb(\xb\vert\zb) = p_\epsb(\xb - \fb(\zb))
\end{equation}
which means that %
the value of $\xb$ can be decomposed as $\xb = \fb(\zb) + \epsb$ where $\epsb$ is an independent noise variable with probability density function $p_\epsb(\epsb)$, i.e. $\epsb$ is independent of $\zb$ or $\fb$. We assume that the function $\fb: \RR^n \rightarrow \RR^d $ is injective; but apart from injectivity it can be an arbitrarily complicated nonlinear function. For the sake of analysis we treat the function $\fb$ itself as a parameter of the model; however in practice we can use flexible function approximators such as neural networks.

We describe the model above with noisy and continuous-valued observations $\xb = \fb(\zb) + \epsb$.\footnote{Equation~\eqref{eq:xcz} can be modified to model discrete variables too, as is detailed in Supplementary Material~\ref{app:discrete}, but that requires a bespoke identifiability theory.} 
However, our identifiability results also apply to non-noisy observations $\xb = \fb(\zb)$, which are a special case of Eq.~\eqref{eq:xcz} where $p_\epsb(\epsb)$ is Gaussian with infinitesimal variance. For these reasons, we can use flow-based generative models ~\citep{dinh2014nice} for $\pt(\xb|\zb)$, while maintaining identifiability.

The prior on the latent variables $\pt(\zb|\ub)$ is assumed to be \textit{conditionally} factorial, where each element of $z_i \in \zb$ has a univariate exponential family distribution given conditioning variable $\ub$. The conditioning on $\ub$ is through an arbitrary function $\lambdab(\ub)$ (such as a look-up table or neural network) that outputs the individual exponential family parameters $\lambda_{i,j}$. The probability density function is thus given by:
\begin{equation}
    \label{eq:zcu}
    p_{\Tb, \lambdab}(\zb\vert\ub) =\prod_i \frac{Q_i(z_i)}{Z_i(\ub)} \exp\left[\sum_{j=1}^k T_{i,j}(z_i) \lambda_{i,j}(\ub)\right]
\end{equation}
where $Q_i$ is the base measure, $Z_i(\ub)$ is the normalizing constant and $\Tb_i=(T_{i,1}, \dots, T_{i,k})$ are the sufficient statistics and $\lambdab_{i}(\ub) = (\lambda_{i,1}(\ub),\dots, \lambda_{i,k}(\ub))$ the corresponding parameters, crucially depending on $\ub$. Finally, $k$, the dimension of each sufficient statistic, is fixed (not estimated). Note that exponential families have universal approximation capabilities, so this assumption is not very restrictive \citep{sriperumbudur2017density}.

\subsection{Estimation by VAE}
\label{model}

Next we propose a practical estimation method for the model introduced above.
Consider we have a dataset $\mathcal{D} = \left\{\left(\xb^{(1)}, \ub^{(1)}\right), \dots, \left(\xb^{(N)}, \ub^{(N)}\right)\right\}$ of observations generated according to the generative model defined in Eq.~\eqref{eq:gen}. 
We propose to use a VAE as a means of learning the true generating parameters $\thetab^*:=(\fb^*, \Tb^*, \lambdab^*)$, up to the indeterminacies discussed below.

VAEs are a framework that simultaneously learns a deep latent generative model and a variational approximation $\qp(\zb\vert\xb,\ub)$ of its true posterior $p_{\thetab}(\zb\vert\xb,\ub)$, the latter being often intractable. Denote by   $\pt(\xb\vert\ub) = \int \pt(\xb, \zb, \vert\ub)\dd\zb$ the conditional marginal distribution of the observations, and with $q_{\mathcal{D}}(\xb, \ub)$ we denote the empirical data distribution given by dataset $\mathcal{D}$. VAEs learn the vector of parameters $(\thetab, \phib)$ by maximizing $\LL(\thetab, \phib)$, a lower bound on the data log-likelihood defined by:
\begin{multline}
\label{eq:loss}
\EE_{q_{\mathcal{D}}}\left[ \log \pt(\xb\vert \ub)\right] \geq \LL (\thetab, \phib) := \\
\EE_{q_{\mathcal{D}}} \left[ \EE_{\qp(\zb|\xb,\ub)} \left[\log\pt(\xb,\zb\vert\ub) - \log \qp(\zb\vert\xb,\ub)\right]\right] 
\end{multline}
We use the reparameterization trick \citep{kingma2013autoencoding} to sample from $\qp(\zb\vert\xb,\ub)$. This trick provides a low-variance stochastic estimator for gradients of the lower bound with respect to $\phib$. The training algorithm is the same as in a regular VAE. Estimates of the latent variables can be obtained by sampling from the variational posterior.

VAEs, like any maximum likelihood estimation method, requires the densities to be normalized. To this end, in practice we choose the prior $p_\thetab(\zb\vert\ub)$ to be a Gaussian location-scale family, which is widely used with VAEs.\footnote{As mentioned in section \ref{sec:model}, our model contains normalizing flows as a special case when $\mathrm{Var}(\eps) = 0$ and the mixing function $\fb$ is parameterized as an invertible flow \citep{rezende2015variational}. Thus, as an alternative estimation method, we could then optimize the log-likelihood directly: 
    $\EE_{q_\DD(\xb, \ub)}[\log p_\theta(\xb\vert\ub)] = \log p_\theta(\fb^{-1}(\zb)\vert\ub) + \log\snorm{J_{\fb^{-1}}(\xb)} $
where $J_{\fb^{-1}}$ is easily computable. The conclusion on consistency given in section \ref{sec:estth} still holds in this case.}

\subsection{Identifiability and consistency results}

As discussed in section~\ref{sec:identifiability}, identifiability as defined by equation~\eqref{eq:identifiable} is very hard to achieve in deep latent variable models. As a first step towards an identifiable model, we seek to recover the model parameters or the latent variables up to trivial transformations. 
Here, we state informally our results on this weaker form of identifiability of the model
---a rigorous treatment is given in Section~\ref{maths.sec}. 
Consider for simplicity the case of no noise and sufficient statistics of size $k=1$, and define $T_i:=T_{i,1}$. Then we can recover $\zb$ which are related to the original $\zb^*$ as follows:
\begin{equation}
    \label{eq:simpleiden}
    (T_{1}^*(z_1^*),\ldots, T_{n}^*(z_n^*))=A (T_{1}(z_1),\ldots, T_{n}(z_n))
\end{equation}
for an invertible matrix $A$.
That is, we can recover the original latent variables up to a component-wise (point-wise) transformations $T_{i}^*, T_{i}$, which are defined as the sufficient statistics of exponential families, and up to a subsequent linear transformation $A$.
Importantly, the linear transformation $A$ can often be resolved by excluding families where, roughly speaking, only the location (mean) is changing. Then $A$ is simply a permutation matrix, and equation \eqref{eq:simpleiden} becomes
\begin{equation}
    \label{eq:simpleiden2}
    T_{i}^*(z_i^*) = T_{{i'}}(z_{i'})
\end{equation}
for a permuted index $i'$.
Thus, the only real indeterminacy is often the component-wise transformations of the latents, which may be inconsequential in many applications.

\subsection{Interpretation as nonlinear ICA}
\label{ica}

Now we show how the model above is closely related to previous work on nonlinear ICA.
In nonlinear ICA, we assume observations $\xb \in \RR^d$, which are the result of an unknown (but invertible) transformation $\mathbf{f}$ of latent variables $\zb \in \RR^d$: \begin{equation}
\label{eq:mixing}
	\xb = \mathbf{f}(\zb)
\end{equation}
where $\zb$ are assumed to follow a factorized (but typically unknown) distribution $p(\zb) = \prod_{i=1}^d p_i(z_i)$. 
This model is essentially a deep generative model. The difference to the definition above is mainly in the lack of noise and the equality of the dimensions: The transformation $\fb$ is deterministic and invertible. Thus, any posteriors would be degenerate.

The goal is then to recover (identify) $\mathbf{f}^{-1}$, which gives the independent components as $\zb = \mathbf{f}^{-1}(\xb)$, based on a dataset of observations of $\xb$ alone. Thus, the goal of nonlinear ICA was always identifiability, which is in general not attained by deep latent variable models, as was discussed in Section~\ref{unidentifiability} above.

To obtain identifiability, we either have to restrict $\fb$ (for instance make it linear) and/or we have to introduce some additional constraints on the distribution of the sources $\zb$. 
Recently, three new nonlinear ICA frameworks \citep{hyvarinen2016unsupervised,hyvarinen2017nonlinear,hyvarinen2019nonlinear} exploring the latter direction were proposed, in which it is possible to recover identifiable sources, up to some trivial transformations. 

The framework in \cite{hyvarinen2019nonlinear} is particularly close to what we proposed above. However, there are several important differences. First, here we define a  generative model where posteriors are non-degenerate, which allows us to show an explicit connection to VAE. We are thus also able to perform maximum likelihood estimation, in terms of evidence lower bound, while previous nonlinear ICA used more heuristic self-supervised schemes. 
Computing  a lower bound on the likelihood is useful, for example, for model selection and validation. In addition, we can in fact prove a tight link between maximum likelihood estimation and maximization of independence of latents, as discussed in Supplementary Material \ref{app:mlmi}. We  also learn both the forward and backward models, which allows for recovering independent latents from data, but also generating new data. The forward model is also likely to help investigate the meaning of the latents. 
At the same time, we are able to provide stronger identifiability results which apply for more general models than earlier theory, and in particular considers the case where the number of latent variables is smaller than the number of observed variables and is corrupted by noise. Given the popularity of VAEs, our current framework should thus be of interest. Further discussion can be found in Supplementary Material \ref{app:prev}.

%% file: sections/4_technical_identifiability.tex
\section{IDENTIFIABILITY THEORY} 
\label{maths.sec}

Now we give our main technical results. The proofs are in Supplementary Material \ref{app:proofs}. 

\paragraph{Notations}
Let $\ZZ\subset\RR^n$ and $\XX\subset\RR^d$ be the domain and the image of $\fb$ in \eqref{eq:xcz}, respectively, and $\UU\subset\RR^m$ the support of the distribution of $\ub$. We denote by $\fb^{-1}$ the inverse defined from $\XX\rightarrow\ZZ$. We suppose that $\ZZ$, $\XX$ and $\UU$ are open sets. We denote by $\Tb(\zb) := \left( \Tb_1(z_1), \dots, \Tb_n(z_n)\right) = \left(T_{1,1}(z_1) \dots, T_{n,k}(z_n)\right) \in \RR^{nk}$ the vector of sufficient statistics of \eqref{eq:zcu}, $\lambdab(\ub) = \left(\lambdab_{1}(\ub), \dots,\lambdab_{n}(\ub)\right) = \left(\lambda_{1,1}(\ub), \dots,\lambda_{n,k}(\ub)\right) \in \RR^{nk}$ the vector of its parameters. Finally $\Theta = \{\thetab := (\fb, \Tb, \lambdab)\}$ is the domain of parameters describing \eqref{eq:gen}. 

\subsection{General results}

In practice, we are often interested in models that are identifiable up to a class of transformation. Thus, we introduce the following definition:

\begin{definition}
Let $\sim$ be an equivalence relation on $\Theta$. We say that \eqref{eq:generalgen} is identifiable up to $\sim$ (or $\sim$-identifiable) if 
\begin{equation}
    p_{\thetab}(\xb) = p_{\tilde{\thetab}}(\xb) \implies \tilde{\thetab} \sim \thetab
\end{equation}
The elements of the quotient space $\bigslant{\Theta}{\sim}$ are called the identifiability classes.
\end{definition}

We now define two equivalence relations on the set of parameters $\Theta$.

\begin{definition}
\label{def:sim}
Let $\sim$ be the equivalence relation
on $\Theta$ defined as follows:
\begin{multline}
    \label{eq:sim}
        (\fb, \Tb, \lambdab) \sim (\tilde{\fb}, \tilde{\Tb}, \tilde{\lambdab}) \Leftrightarrow \\ \exists A, \mathbf{c} \mid \Tb(\fb^{-1}(\xb)) = A \tilde{\Tb}(\tilde{\fb}^{-1}(\xb)) + \mathbf{c},\forall \xb \in \XX
\end{multline} 
where $A$ is an $nk\times nk$ matrix and $\mathbf{c}$ is a vector

If $A$ is \emph{invertible}, we denote this relation by $\sim_A$. If $A$ is a block \emph{permutation}\footnote{each block linearly transforms $\Tb_i$ into $\tilde{\Tb}_{i'}$.}  matrix, we denote it by~$\sim_P$.
\end{definition}

Our main result is the following Theorem\footnote{an alternative version is in Supplementary Material \ref{app:altiden}.}:
\begin{theorem}
\label{th:iden}
Assume that we observe data sampled from a generative model defined according to \eqref{eq:gen}-\eqref{eq:zcu}, with parameters $(\fb, \Tb, \lambdab)$. Assume the following holds:

\begin{enumerate}[label=(\roman*)]
    \item \label{th:iden:ass1} The set $\{\xb \in \XX \vert \varphi_\eps (\xb) = 0\}$ has measure zero, where $\varphi_\eps$ is the characteristic function of the density $p_\eps$ defined in \eqref{eq:xcz}.
    
    \item \label{th:iden:ass2} The mixing function $\fb$ in \eqref{eq:xcz} is injective.

    \item \label{th:iden:ass3} The sufficient statistics $T_{i,j}$ in \eqref{eq:zcu} are differentiable almost everywhere, and $(T_{i,j})_{1\leq j \leq k}$ are linearly independent on any subset of $\XX$ of measure greater than zero.
    
	\item \label{th:iden:ass4} There exist $nk+1$ distinct points $\ub^0, \dots, \ub^{nk}$ such that the matrix
		\begin{equation}
		\label{eq:L}
    		L = \left( \lambdab(\ub_1)-\lambdab(\ub_0), \dots, \lambdab(\ub_{nk}) - \lambdab(\ub_0) \right)
		\end{equation}
		 of size $nk \times nk$ is invertible.\footnote{the intuition and feasibility of this assumption are discussed in Supplementary Material \ref{app:understanding}.}

\end{enumerate}
then the parameters $(\fb, \Tb, \lambdab)$ are $\sim_A$-identifiable.
\end{theorem}

This Theorem guarantees a basic form of identifiability of the generative model \eqref{eq:gen}. In fact, suppose the data was generated according to the set of parameters $(\fb, \Tb, \lambdab)$. And let $(\tilde{\fb}, \tilde{\Tb}, \tilde{\lambdab})$ be the parameters obtained from some learning algorithm (supposed consistent in the limit of infinite data) that perfectly approximates the marginal distribution of the observations. Then the Theorem says that necessarily $(\tilde{\fb}, \tilde{\Tb}, \tilde{\lambdab}) \sim_A (\fb, \Tb, \lambdab)$. If there were no noise, this would mean that the learned transformation $\tilde{\fb}$ transforms the observations into latents $\tilde{\zb} = \tilde{\fb}^{-1}(\xb)$ that are equal to the true generative latents $\zb = \fb^{-1}(\xb)$, up to a linear invertible transformation (the matrix $A$) and point-wise nonlinearities (in the form of $\Tb$ and $\tilde{\Tb}$). With noise, we obtain the posteriors of the latents up to an analogous indeterminacy. 

\subsection{Characterization of the linear indeterminacy}

The equivalence relation $\sim_A$ provides a useful form of identifiability, but it is very desirable to remove the linear indeterminacy $A$, and reduce the equivalence relation to $\sim_P$ by analogy with linear ICA where such matrix is resolved up to a \emph{permutation} and \emph{signed scaling}. We present in this section sufficient conditions for such reduction, and special cases to avoid.

We will start by giving two Theorems that provide sufficient conditions. Theorem \ref{th:iden2} deals with the more general case $k\geq 2$, while Theorem \ref{th:iden3} deals with the special case $k=1$.

\begin{theorem}[$k\geq 2$]
\label{th:iden2}
Assume the hypotheses of Theorem \ref{th:iden} hold, and that $k\geq 2$. Further assume:
\begin{enumerate}[label=(2.\roman*), leftmargin=1cm]
    \item \label{th:iden2:ass2} The sufficient statistics $T_{i,j}$ in \eqref{eq:zcu} are twice differentiable.
    \item \label{th:iden2:ass4} The mixing function $\fb$ has all second order cross derivatives.
\end{enumerate}
then the parameters $(\fb, \Tb, \lambdab)$ are $\sim_P$-identifiable.
\end{theorem}

\begin{theorem}[$k=1$]
\label{th:iden3}
Assume the hypotheses of Theorem \ref{th:iden} hold, and that $k=1$. Further assume:

\begin{enumerate}[label=(3.\roman*), leftmargin=1cm]
    \item \label{th:iden3:ass2} The sufficient statistics $T_{i,1}$ are not monotonic\footnote{monotonic means it is strictly increasing or decreasing.}.
    \item \label{th:iden3:ass4} All partial derivatives of $\fb$ are continuous.
\end{enumerate}
then the parameters $(\fb, \Tb, \lambdab)$ are $\sim_P$-identifiable.
\end{theorem}

These two Theorems imply that in most cases $\tilde{\fb}^{-1}\circ\fb:\ZZ\rightarrow\ZZ$ is a pointwise
\footnote{each of its component is a function of only one $z_i$.}
nonlinearity, which essentially means that the estimated latent variables $\tilde{\zb}$ are equal to a permutation and a pointwise nonlinearity of the original latents $\zb$. 

This kind of identifiability is stronger than any previous results in the literature, and considered sufficient in many applications. 
On the other hand, there are very special cases where a linear indeterminacy cannot be resolved, as shown by the following:

\begin{proposition}
\label{prop:nec}
Assume that $k=1$, and that
\begin{enumerate}[label=(\roman*)]
    \item $T_{i,1}(z_i)=z_i$ for all $i$.
    \item $Q_i(z_i) = 1$ or $Q_i(z_i) = e^{-z_i^2}$ for all $i$.
\end{enumerate} 
Then $A$ can not be reduced to a permutation matrix.
\end{proposition}

This Proposition stipulates that if the components are Gaussian (or exponential in the case of non-negative components) and \textit{only} the location is changing, we can't hope to reduce the matrix $A$ in $\sim_A$ to a permutation. In fact, to prove this in the Gaussian case, we simply consider orthogonal transformations of the latent variables, which all give rise to the same observational distribution with a simple adjustment of parameters.

\subsection{Consistency of Estimation}
\label{sec:estth}

The theory above further implies a consistency result on the VAE. If the variational distribution $\qp$ is a broad parametric family that includes the true posterior, then we have the following result.

\begin{theorem}
\label{th:vaeiden}
Assume the following:
\begin{enumerate}[label=(\roman*)]
    \item The family of distributions $\qp(\zb\vert\xb,\ub)$ contains $\pftl(\zb\vert\xb,\ub)$.
    \item We maximize $\LL(\thetab, \phib)$ with respect to both $\thetab$ and~$\phib$.
\end{enumerate}
    then in the limit of infinite data, the VAE learns the true parameters $\thetab^*:=(\fb^*, \Tb^*, \lambdab^*)$ up to the equivalence class defined by $\sim$ in \eqref{eq:sim}.
\end{theorem}

%% file: sections/5_experiments.tex
\section{EXPERIMENTS}

\subsection{Simulations on artifical data}
\label{simulation}

\paragraph{Dataset} 
We run simulations on data used previously in the nonlinear ICA literature \citep{hyvarinen2016unsupervised, hyvarinen2019nonlinear}. 
We generate synthetic datasets where the sources are non-stationary Gaussian time-series: we divide the sources into $M$ segments of $L$ samples each. The conditioning variable $\ub$ is the segment label, and its distribution is uniform on the integer set $\iset{1,M}$. Within each segment, the conditional prior distribution is chosen from the family \eqref{eq:zcu} for small $k$. When $k=2$, we used mean and variance modulated Gaussian distribution. When $k=1$, we used variance modulated Gaussian or Laplace (to fall within the hypotheses of Theorem \ref{th:iden3}). The true parameters $\lambda_i$ were randomly and independently generated across the segments and the components from a non degenerate distributions to satisfy assumption \ref{th:iden:ass4} of Theorem \ref{th:iden}. 
Following \cite{hyvarinen2019nonlinear}, we  mix the sources using a multi-layer perceptron (MLP) and add small Gaussian noise.  

\paragraph{Model specification} %
Our estimates of the latent variables are generated from the variational posterior $\qp(\zb\vert\ub,\xb)$, for which we chose the following form: $\qp(\zb\vert\xb,\ub) = \mathcal{N}\left(\zb|\gb(\xb, \ub; \phi_\gb), \diag{\sigmab^2(\xb, \ub; \phi_\sigmab)}\right)$, a multivariate Gaussian with a diagonal covariance. The noise distribution $p_\eps$ is Gaussian with small variance.
The functional parameters of the decoder and the inference model, as well as the conditional prior are chosen to be MLPs. We use an Adam optimizer \citep{kingma2014adam} to update the parameters of the network by maximizing $\LL(\thetab, \phib)$ in equation~\eqref{eq:loss}. The data generation process as well as  hyperparameter choices are detailed in Supplementary Material \ref{app:model_details}.

\paragraph{Performance metric}
To evaluate the performance of the method, we compute the mean correlation coefficient (MCC) between the original sources and the corresponding latents sampled from the learned posterior. 
To compute this performance metric, we first calculate all pairs of correlation coefficients between source and latent components. We then solve a linear sum assignment problem to assign each latent component to the source component that best correlates with it, thus reversing any permutations in the latent space. 
A high MCC means that we successfully identified the true parameters and recovered the true sources, up to point-wise transformations. This is a standard measure used in ICA.

\begin{figure}
\begin{subfigure}{.12\textwidth}
  \center{\includegraphics[width=\textwidth]
    {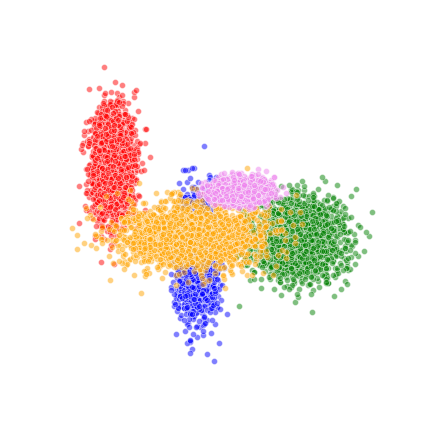}}
    \caption{\label{fig:tcl_2d_s} $p_{\thetab^*}(\zb\vert\ub)$}
\end{subfigure}%
\begin{subfigure}{.12\textwidth}
  \center{\includegraphics[width=\textwidth]
    {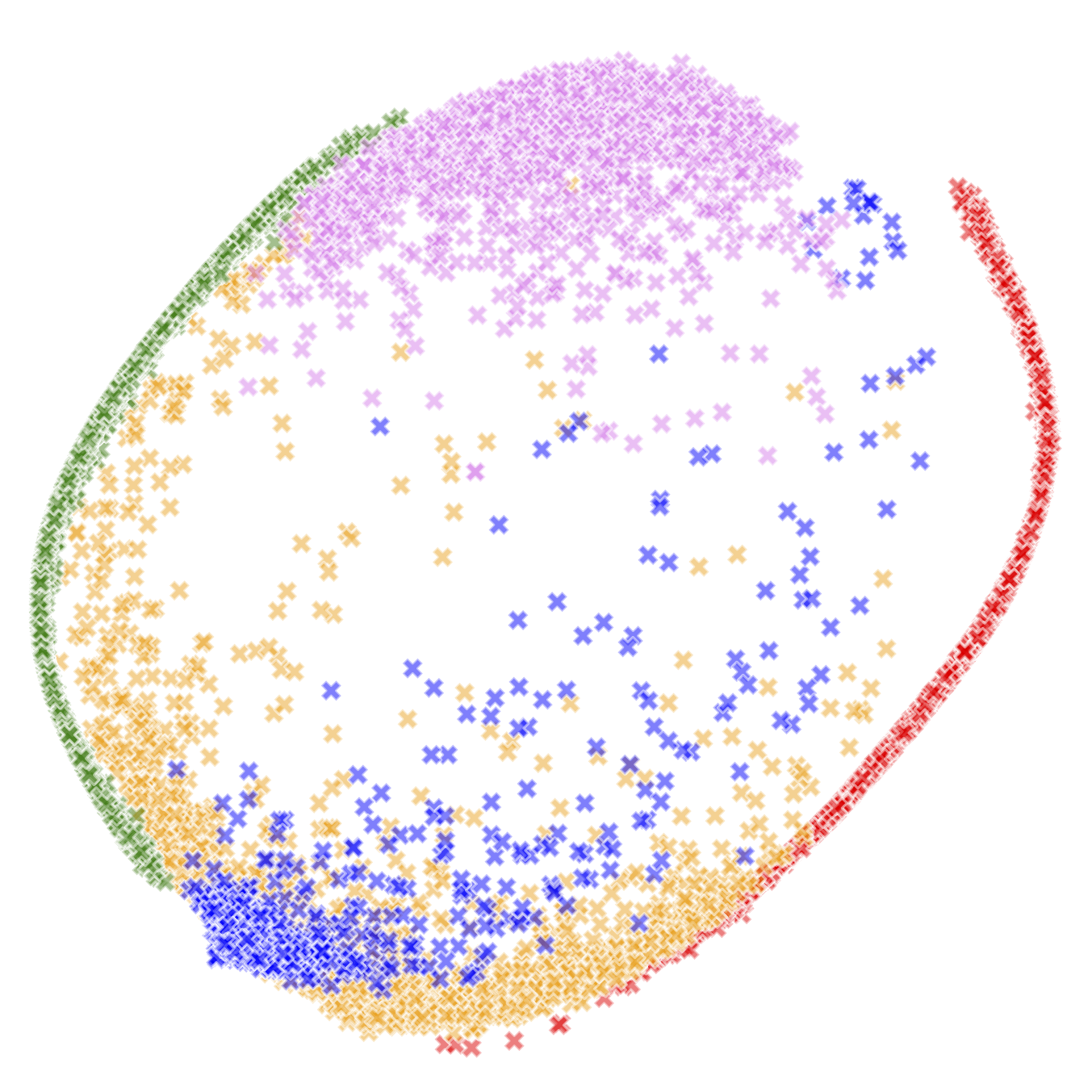}}
    \caption{\label{fig:tcl_2d_z}$p_{\thetab^*}(\xb\vert\ub)$}
\end{subfigure}%
\begin{subfigure}{.12\textwidth}
  \center{\includegraphics[width=\textwidth]
    {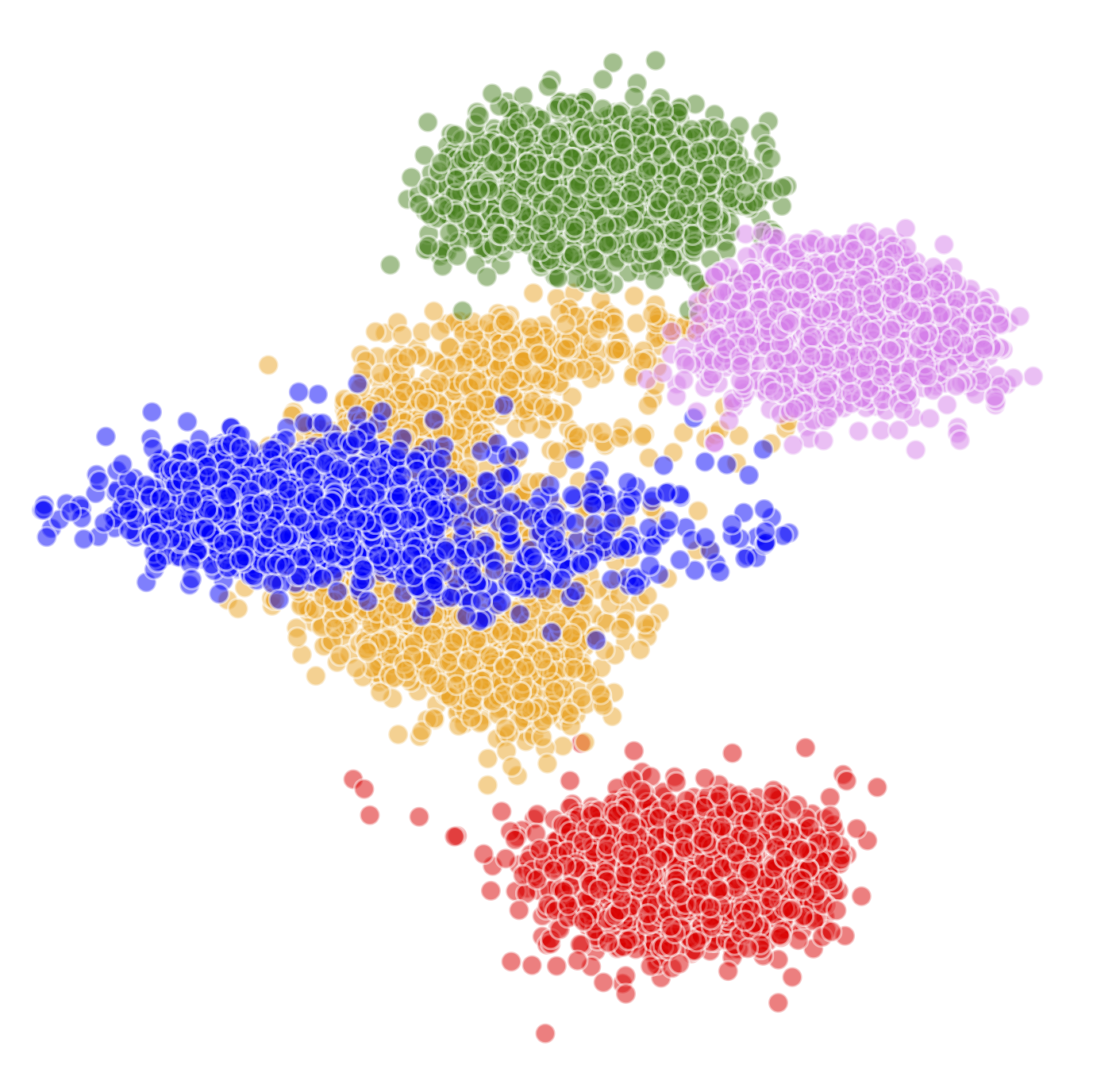}}
    \caption{\label{fig:tcl_2d_f}$p_\thetab(\zb\vert\xb,\ub)$}
\end{subfigure}%
\begin{subfigure}{.12\textwidth}
  \center{\includegraphics[width=\textwidth]
    {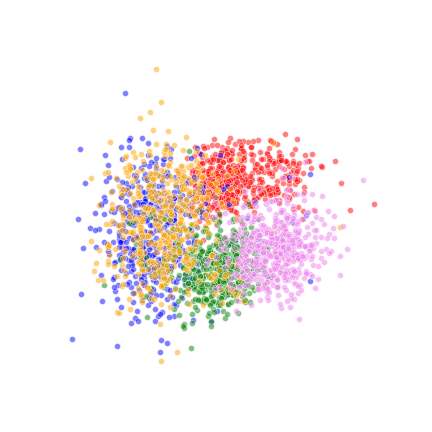}}
    \caption{\label{fig:tcl_2d_vae}$p_{\textrm{VAE}}(\zb\vert\xb)$}
\end{subfigure}
\caption{\label{fig:tcl_2d} Visualization of both observation and latent spaces in the case $n=d=2$ and where the number of segments is $M=5$ (segments are colour coded). First, data is generated in (a)-(b) as follows: $(a)$ samples from the true distribution of the sources $p_{\thetab^*}(\zb\vert\ub)$: Gaussian with non stationary mean and variance, $(b)$ are observations sampled from $p_{\thetab^*}(\xb\vert\zb)$. Second, after learning both a vanilla VAE and an iVAE models, we plot in $(c)$ the latent variables sampled from the posterior $\qp(\zb\vert\xb, \ub)$ of the iVAE and in $(d)$ the latent variables sampled from the posterior of the vanilla VAE.
}
\end{figure}

\paragraph{Results: 2D example}
First, we show a visualization of identifiability of \VAEICA in a 2D case in Figure \ref{fig:tcl_2d}, where we plot the original sources, observed data and the posterior distributions learned by our model, compared to a vanilla VAE. Our method recovers the original sources up to trivial indeterminacies (rotation and sign flip), whereas the VAE fails to do a good separation of the latent variables.

\begin{figure}
\begin{subfigure}{.24\textwidth}
  \center{\includegraphics[width=\textwidth]
    {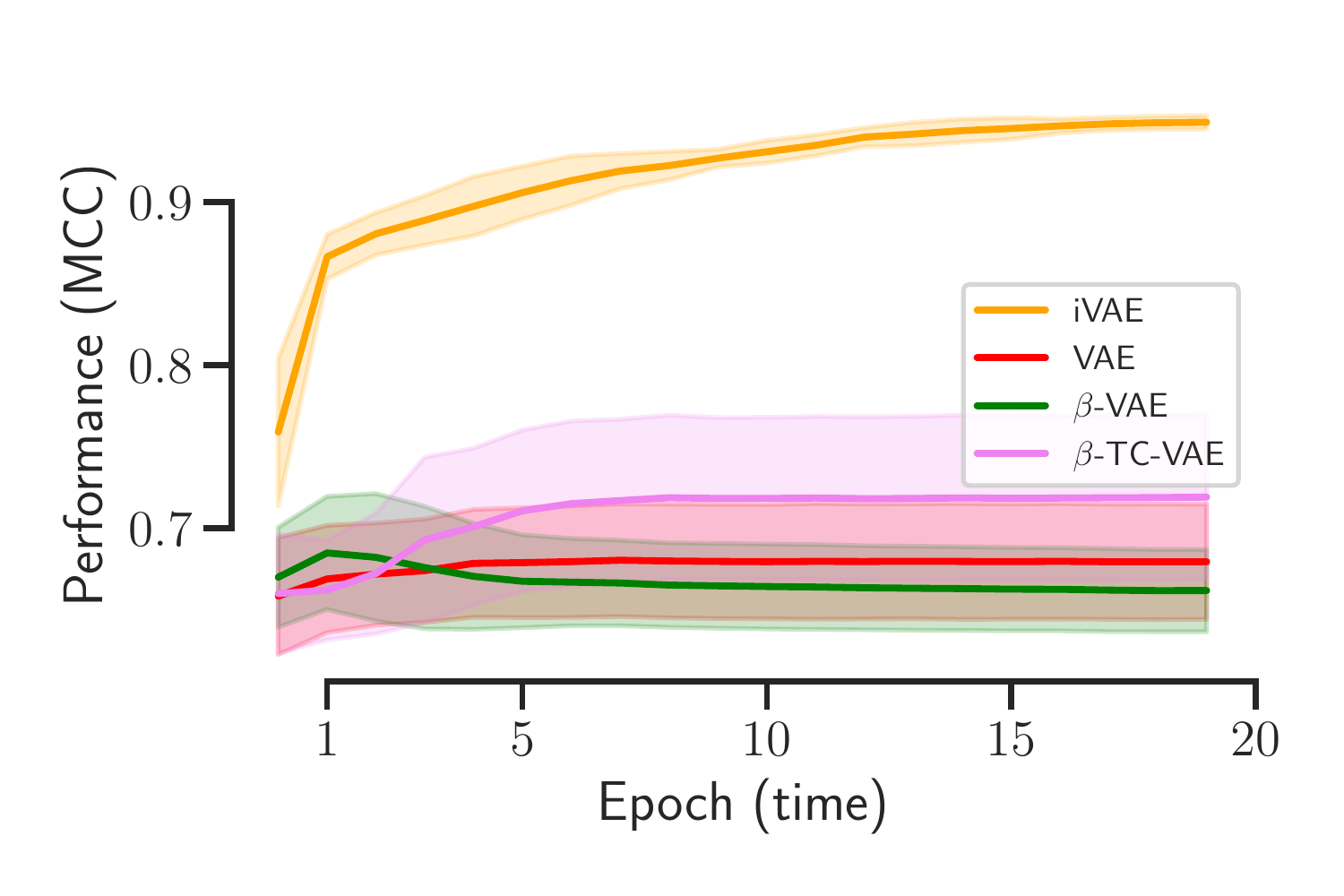}}
    \caption{\label{fig:tcl_perf} Training dynamics}
\end{subfigure}%
\begin{subfigure}{.24\textwidth}
  \center{\includegraphics[width=\textwidth]
    {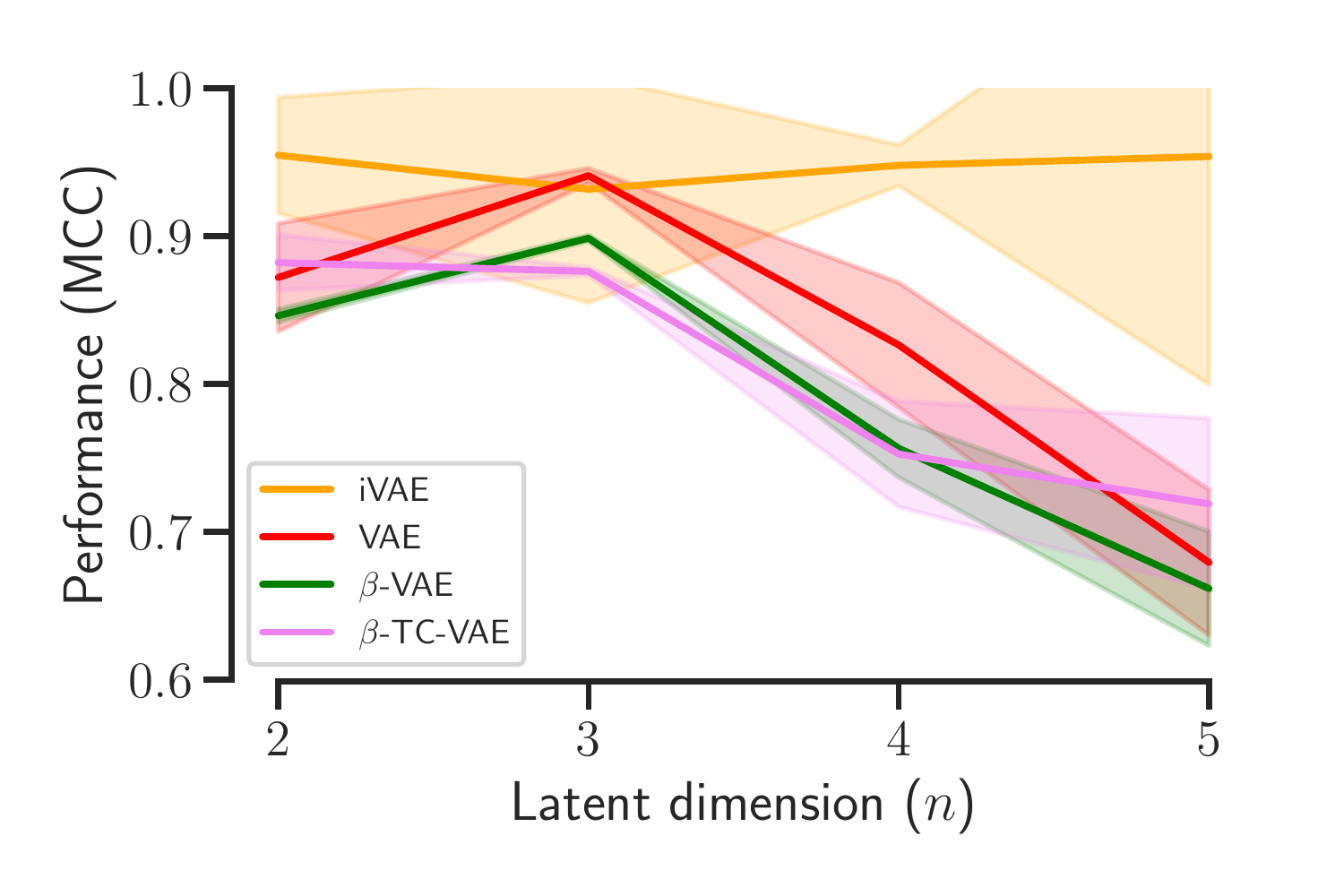}}
    \caption{\label{fig:tcl_dim} Changing $n$}
\end{subfigure}
\caption{\label{fig:tcl_comp} Performance of \VAEICA in recovering the true sources, compared to VAE, $\beta$-VAE and $\beta$-TC-VAE, for $M=40$, $L=1000$ and $d=5$ (and $n=5$ for $(a)$).}
\end{figure}

\paragraph{Results: Comparison to VAE variants}
We compared the performance of \VAEICA to a vanilla VAE. We used the same network architecture for both models, with the sole exception of the addition of the conditional prior in \VAEICA. When the data is centered, the VAE prior is Gaussian or Laplace.  We also compared the performance to two models from the disentanglement literature, namely a $\beta$-VAE \citep{higgins2016betavae} and a $\beta$-TC-VAE \citep{chen2018isolating}. 
The parameter $\beta$ of the $\beta$-VAE and the parameters $\alpha$, $\beta$ and $\gamma$ for $\beta$-TC-VAE were chosen by following the instructions of their respective authors.
We trained these 4 models on the dataset described above, with $M=40$, $L=1000$, $d=5$ and $n \in [2, 5]$.
Figure \ref{fig:tcl_perf} compares performances obtained from an optimal choice of parameters achieved by \VAEICA and the three models discussed above, when the dimension of the latent space equals the dimension of the data ($n=d=5$). \VAEICA achieved an MCC score of above $95\%$, whereas the other three models fail at finding a good estimation of the true parameters. 
We further investigated the impact of the latent dimension on the performance in  Figure \ref{fig:tcl_dim}. \VAEICA has much higher correlations than the three other models, especially as the dimension increases. Further visualization are in Supplementary Material \ref{app:figs}.

\begin{figure}
\begin{subfigure}{.24\textwidth}
    \center{\includegraphics[width=\textwidth]
    {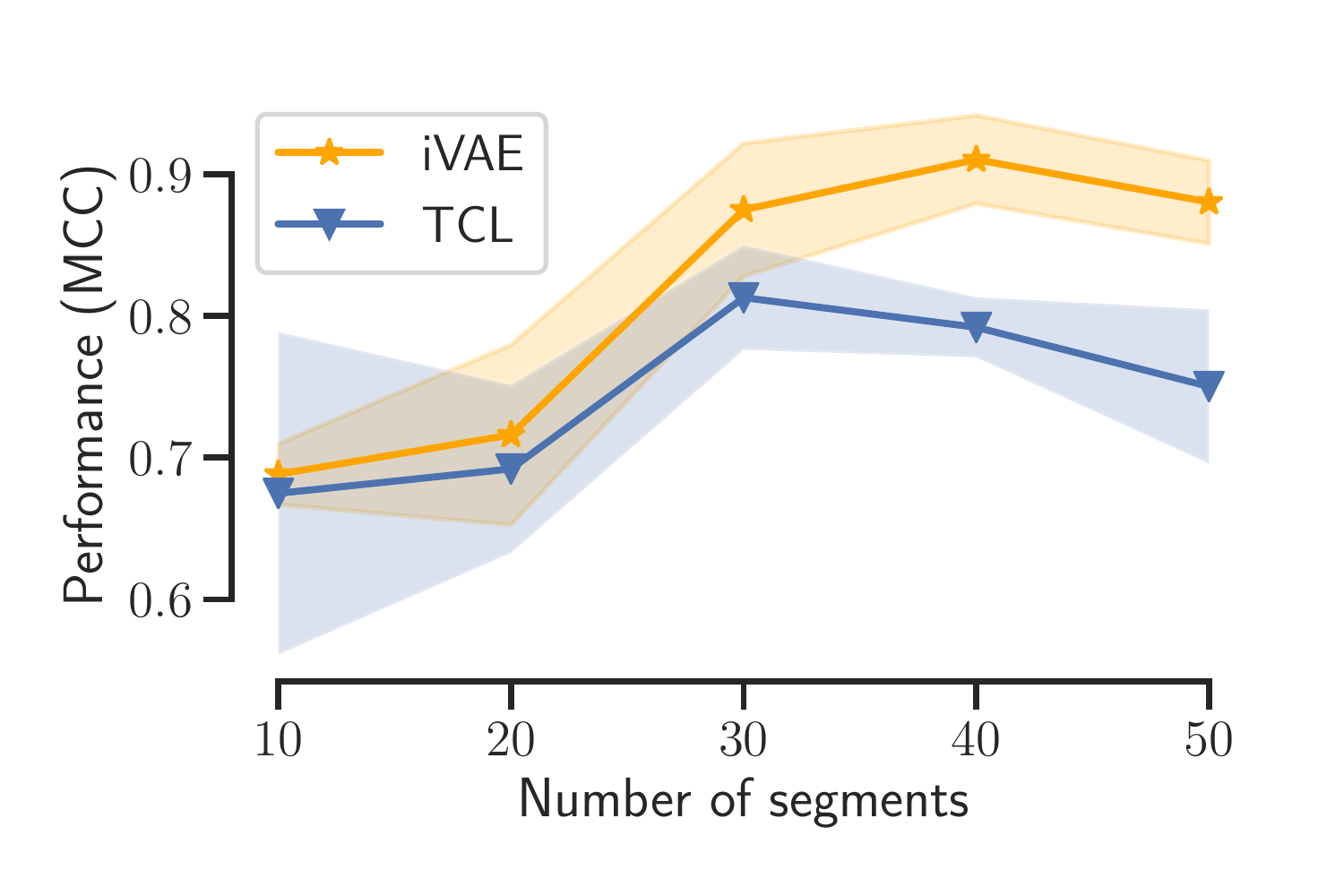}}
    \caption{\label{fig:vae_vs_tcl_normal} Normal}
\end{subfigure}%
\begin{subfigure}{.24\textwidth}
    \center{\includegraphics[width=\textwidth]
    {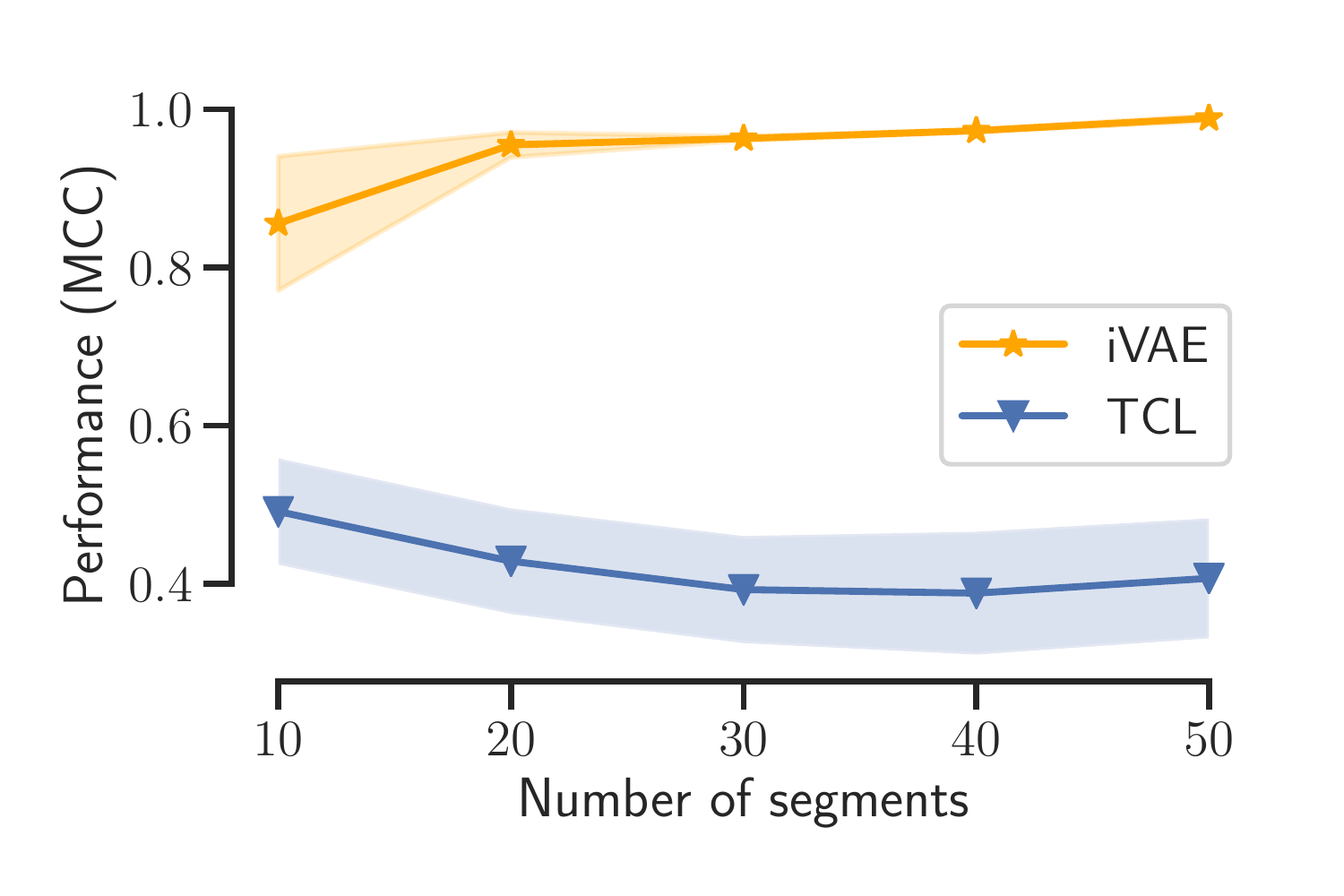}}
    \caption{\label{fig:vae_vs_tcl_staircase} Easy to classify}
\end{subfigure} 
\caption{\label{fig:vae_vs_tcl} $(a)$ Performance of iVAE in comparison to TCL in recovering the true sources on normal data $(b)$ Performance of iVAE in comparison to TCL in recovering the true sources on easy to classify data.}
\end{figure}

\paragraph{Results: Comparison to TCL}
Next, we compared our method to previous nonlinear ICA methods, namely TCL by \cite{hyvarinen2016unsupervised}, which is based on a self supervised classification task (see Supplementary Material \ref{app:prev_ica}). We run simulations on the same dataset as Figure \ref{fig:tcl_perf}, where we varied the number of segments from 10 to 50. 
Our method slightly outperformed TCL in our experiments. The results are reported in Figure \ref{fig:vae_vs_tcl_normal}. Note that according to \citet{hyvarinen2019nonlinear}, TCL performs best among previously proposed methods for this kind of data.

Finally, we wanted to show that our method is robust to some failure modes which occur in the context of self-supervised methods. The theory of TCL is premised on the notion that in order to accurately classify observations into their relative segments, the model must learn the true log-densities of sources within each segment. While such theory will hold in the limit of infinite data, we considered here a special case where accurate classification did not require learning the log-densities very precisely. This was achieved by generating synthetic data where $x_2$ alone contained sufficient information to perform classification, by  making the mean 
of $x_2$ significantly modulated across segments; 
further details in 
Supplementary Material \ref{app:easy_data}.
In such a setting, TCL is able to obtain high classification accuracy without unmixing observations, resulting in its failure to recover latent variables as reflected in Figure \ref{fig:vae_vs_tcl_staircase}. In contrast, the proposed iVAE, by virtue of optimizing a maximum likelihood objective, does not suffer from such degenerate behaviour. 

\textbf{Further simulations} on  hyperparameter selection and discrete data are in Supplementary Material~\ref{app:extra_exp}.

\subsection{Nonlinear causal discovery in fMRI}
\label{sec:realdata}

An important application of %
ICA methods is within the domain of causal discovery
\citep{peters2017elements}. 
The use of ICA methods in this domain is premised on the equivalence between a (nonlinear) ICA model and 
the corresponding structural equation model (SEM). Such a connection was 
initially exploited in the linear case \citep{shimizu2006linear}
and  extended to the nonlinear case 
by \cite{monti2019causal} who employed TCL. 

Briefly, consider
data $\xb=(x_1, x_2)$. 
The goal is to establish if the causal direction is $x_1 \rightarrow x_2$, or
$x_2 \rightarrow x_1$,  
or conclude that no (acyclic) causal relationship exists. 
Assuming %
$x_1 \rightarrow x_2$, then the problem can be described by the following SEM:
$    x_1 = f_1(n_1), ~x_2 = f_2(x_1, n_2)$
where $\fb=(f_1, f_2)$ is a (possibly nonlinear) mapping and $\mathbf{n} = (n_1, n_2)$ are latent disturbances that are assumed to be independent.
The above SEM can be seen as a nonlinear ICA model where 
latent disturbances,
$\mathbf{n}$, are the sources. %
As such, we may perform causal discovery by first recovering latent 
disturbances (using TCL or iVAE) and then running a series of 
independence tests. Formally, if
$x_1 \rightarrow x_2$ then, denoting statistical independence by $\independent$, it suffices to verify that %
$x_1 \independent n_2$
whereas $x_1 \notindependent n_1$, $x_2 \notindependent n_1$ and $x_2 \notindependent n_2$. %
Such an approach can be extended beyond two-dimensional 
observations as described in \cite{monti2019causal}.

\begin{figure}
\begin{subfigure}{.24\textwidth}
    \center{\includegraphics[width=\textwidth]
    {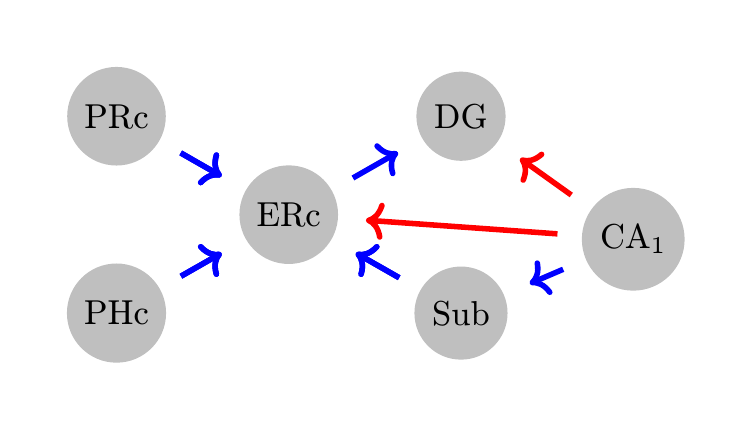}}
    \caption{\label{fig:fmri_iVAE} iVAE}
\end{subfigure}%
\begin{subfigure}{.24\textwidth}
    \center{\includegraphics[width=.9\textwidth]
    {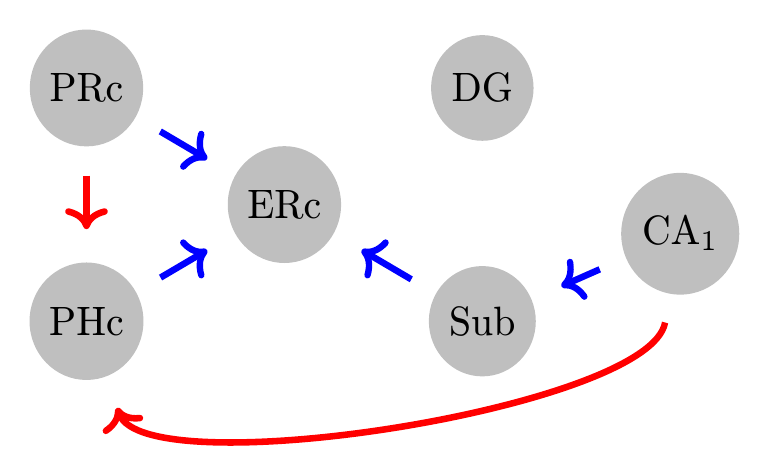}}
    \caption{\label{fig:smri_TCL} TCL}
\end{subfigure}
\caption{\label{fig:fmri_nonsens} Estimated causal graph on hippocampal fMRI data unmixing of sources is achieved 
via iVAE (left) or TCL (right). Blue edges are feasible 
given anatomical connectivity, red edges are not. 
}
\end{figure}

To demonstrate the benefits of iVAE as compared to TCL, both algorithms 
were employed to learn causal structure from 
fMRI data (details in Supplementary Material \ref{app:fmri}).
The recovered causal graphs are shown in Figure \ref{fig:fmri_nonsens}.
Blue edges
are anatomically feasible whilst red edges are not.
There is significant overlap between the estimated causal networks, but
in the case of iVAE both anatomically incorrect edges 
correspond to indirect causal effects. This is in contrast with TCL where incorrect edges are incompatible with anatomical structure and cannot be explained as indirect effects.

%% file: sections/6_conclusion.tex
\section{CONCLUSION}\label{sec:conclusion}
Unsupervised learning can have many different goals, such as: $(i)$ approximate the data distribution, $(ii)$ generate new samples, $(iii)$ learn useful features, and above all $(iv)$ learn the original latent code that generated the data (identifiability). Deep latent-variable models typically implemented by VAEs are an excellent framework to achieve $(i)$, and are thus our first building block. The nonlinear ICA model discussed in section \ref{ica} is the only existing framework to provably achieve $(iv)$.  We bring these two pieces together to create our new model termed \VAEICA. In particular, this is the first rigorous proof of identifiability in the context of VAEs. Our model in fact checks all the four boxes above that are desired in unsupervised learning.

The advantage of the new framework over typical deep latent-variable models used with VAEs is that we actually recover the original latents, thus providing principled disentanglement.
On the other hand, the advantages of this algorithm for solving nonlinear ICA over \cite{hyvarinen2019nonlinear} are several; briefly, we significantly strengthen the identifiability results, we obtain the likelihood and can use MLE, we learn a forward model as well and can generate new data, and we consider the more general cases of noisy data with fewer components.

\section*{Corrigendum}

The published version of this paper claimed that the identifiability result extends to the case with discrete observations. Unfortunately, after publication, we found an error in the proof for the discrete case, so we can no longer claim that this is the case. We suspect that the discrete case requires a different type of proof, which we leave for future work. To remove incorrect claims, we made minimal changes to the abstract, Section \ref{sec:model} and Section \ref{sec:conclusion}; we also rewrote Appendix ~\ref{app:discrete} which contained the erroneous proof. We do still provide experiments in Supplementary Material~\ref{app:extra_exp} that strongly suggest that identifiability is achievable in such setting.

%% file: sections/7_appendix_proofs.tex
\clearpage
\onecolumn

\appendix

  \begin{center}
  \textit{\large Supplementary Material for}\\ \ \\
      {\large \bf Variational Autoencoders and Nonlinear ICA: A Unifying Framework}\\ \ \\published at AISTATS 2020
  \end{center}

\section{LEMMAS FOR THE EXPONENTIAL FAMILIES}

We only consider univariate distributions in this section. The domain of the distributions is assumed to be $\RR$, but all results hold if we replace $\RR$ by an open set $\ZZ \subset \RR$ whose Lebesgue measure is greater than $0$.

\subsection{Exponential family distributions}
\begin{definition}[Exponential family]
    A univariate exponential family is a set of distributions whose probability density function can be written as
    \begin{equation}
        \label{eq:exp_fam}
        p(x) = Q(x)Z(\thetab)e^{\dotprod{\Tb(x)}{\thetab}}
    \end{equation}
    where $\Tb:\RR\rightarrow\RR^k$ is called the sufficient statistic, $\thetab\in\RR^k$ is the natural parameter, $Q:\RR\rightarrow\RR$ the base measure and $Z(\thetab)$ the normalization constant. The dimension $k\in\mathbb{N}\setminus\{0\}$ of the parameter is always considered to be \emph{minimal}, meaning that we can't rewrite the density $p$ to have the form \eqref{eq:exp_fam} with a smaller $k'<k$. We call $k$ the size of $p$.
\end{definition}

\begin{lemma}
\label{lem:suffstat_lin2}
Consider an exponential family distribution with $k\geq 2$ components. If there exists $\alpha\in\RR^k$ such that $T_k(x) = \sum_{i=1}^{k-1}\alpha_i T_i(x) + \alpha_k$, then $\alpha=0$. In particular, the components of the sufficient statistic $\Tb$ are linearly independent.
\end{lemma}
\textit{Proof:} Suppose the components $(T_1, \dots, T_k)$ are not linearly independent. Then $\exists \alphab\in\RR^k\setminus\{0\}$ such that $\forall x\in\RR, \; \sum_{i=1}^k \alpha_iT_i(x) = 0$. Suppose $\alpha_k\neq0$ (up to rearrangement of the indices), then we can write $T_k$ as a function of the remaining $T_i, i<k$, contradicting the minimality of $k$. \QED

\subsection{Strongly exponential distributions}

\begin{definition}[Strongly exponential distributions]
\label{def:str_exp}
    We say that an exponential family distribution is \emph{strongly exponential} if for any subset $\XX$ of $\RR$ the following is true: 
    \begin{equation}
    \label{eq:str_exp}
        \left(\exists \, \thetab\in\RR^k \,\vert \, \forall x\in\XX, \dotprod{\Tb(x)}{\thetab} = \textrm{const}\right) 
         \implies \left(l(\XX)=0 \textrm{ or } \thetab = 0\right)
    \end{equation}
    where $l$ is the Lebesgue measure.
\end{definition}
In other words, the density of a strongly exponential distribution has almost surely the exponential component in its expression and can only be reduced to the base measure on a set of measure zero.

\begin{example}
The strongly exponential condition is very general, and is satisfied by all the usual exponential family distributions like the Gaussian, Laplace, Pareto, Chi-squared, Gamma, Beta, \etc
\end{example}

We will now give useful Lemmas that will be used in the proofs of the technical Theorems. 

\begin{lemma}
\label{lem:expfam_app}
Consider a \emph{strongly exponential} family distribution such that its sufficient statistic $\Tb$ is \emph{differentiable} almost surely. Then $T_i' \neq 0$ almost everywhere on $\RR$ for all $1\leq i \leq k$.  
\end{lemma}
\textit{Proof:}
Suppose that $p$ is strongly exponential, and let $\XX=\cup_i\{x\in\RR, T_i'(x) \neq 0\}$. Chose any $\thetab \in \RR^k\setminus\{0\}$. Then $\forall x\in\XX, \dotprod{\Tb'(x)}{\thetab} = 0$. By integrating, we find that $\dotprod{\Tb(x)}{\thetab}=\textrm{const}$. By hypothesis, this means that $l(\XX)=0$.\QED

\begin{lemma}
\label{lem:expfam}
    Consider a strongly exponential distribution of size $k\geq 2$ with sufficient statistic $\Tb(x) = (T_1(x), \dots, T_k(x))$. Further assume that $\Tb$ is \emph{differentiable} almost everywhere. Then there exist $k$ distinct values $x_1$ to $x_k$ such that $(\Tb'(x_1), \dots, \Tb'(x_k))$ are linearly independent in $\RR^k$.
\end{lemma}

\textit{Proof:}
Suppose that for any choice of such $k$ points, the family $(\Tb'(x_1), \dots, \Tb'(x_k))$ is never linearly independent. That means that $\Tb'(\RR)$ is included in a subspace of $\RR^k$ of dimension at most $k-1$. Let $\thetab$ a non zero vector that is orthogonal to $\Tb'(\RR)$. Then for all $x\in\RR$, we have $\dotprod{\Tb'(x)}{\thetab} = 0$. By integrating we find that $\dotprod{\Tb(x)}{\thetab} = \textrm{const}$. Since this is true for all $x\in\RR$ and for a $\thetab\neq0$, we conclude that the distribution is not strongly exponential, which contradicts our hypothesis. \QED

\begin{lemma}
    \label{lem:str_exp_lem}
    Consider a strongly exponential distribution of size $k\geq 2$ with sufficient statistic $\Tb$. Further assume that $\Tb$ is \emph{twice differentiable} almost everywhere. Then
    \begin{equation} 
    \label{eq:str_exp_lem}
        \dim\left(\Span\left(\left(T_i'(x), T_i''(x)\right)^T, 1\leq i \leq k\right)\right) \geq 2 
    \end{equation}
    almost everywhere on $\RR$.
\end{lemma}
\textit{Proof:}
Suppose there exists a set $\XX$ of measure greater than zero where \eqref{eq:str_exp_lem} doesn't hold. This means that the vectors $[T_i'(x), T_i''(x)]^T$ are collinear for any $i$ and for all $x\in\XX$. In particular, it means that there exists $\alpha\in\RR^k\setminus\{0\}$ s.t. $\sum_i\alpha_i T_i'(x) = 0$. By integrating, we get $\dotprod{\Tb(x)}{\alphab} = \textrm{const},\;\forall x \in \XX $. Since $l(\XX)>0$, this contradicts equation $\eqref{eq:str_exp}$. \QED

\begin{lemma}
\label{lem:var_exp_fam}
    Consider $n$ strongly exponential distributions of size $k\geq 2$ with respective sufficient statistics $\Tb_j = (T_{j,1},\dots T_{j,k})$, $1\leq j \leq n$. Further assume that the sufficient statistics are \emph{twice differentiable}. Define the vectors $\eb^{(j,i)} \in \RR^{2n}$, such that $\eb^{(j,i)} = \left(0, \dots, 0, T_{j,i}', T_{j,i}'', 0, \dots, 0\right)$, where the non-zero entries are at indices $(2j, 2j+1)$. Let $\xb:=(x_1, \dots, x_n)\in\RR^n$ Then the matrix $\overline{e}(\xb):=(\eb^{(1,1)}(x_1), \dots, \eb^{(1,k)}(x_1), \dots \eb^{(n,1)}(x_n), \dots, \eb^{(n,k)}(x_n))$ of size $(2n\times nk)$ has rank $2n$ almost everywhere on $\RR^n$.
\end{lemma}
\textit{Proof:}
It is easy to see that the matrix $\overline{e}(\xb)$ has at least rank $n$, because by varying the index $j$ in $\eb^{(j,i)}$ we change the position of the non-zero entries. By changing the index $i$, we change the component within the same sufficient statistic. Now fix $j$ and consider the submatrix $\left[\eb^{(j,1)}(x_j), \dots, \eb^{(j,k)}(x_j)\right]$. By using Lemma \ref{lem:str_exp_lem}, we deduce that this submatrix has rank greater or equal to 2 because its columns span a subspace of dimensions greater or equal to 2 almost everywhere on $\RR$. 
Thus, we conclude that the rank of $\overline{e}(\xb)$ is $2n$ almost everywhere on $\RR^n$.\QED

We will give now an example of an exponential family distribution that is not strongly exponential. 
\begin{example}
Consider an exponential family distribution with density function
\begin{equation}
    p(x) = e^{-x^2}Z(\thetab)\exp\left(\theta_1\min(0,x) - \theta_2\max(0,x)\right)
\end{equation}
This density sums to 1 and $Z(\thetab)$ is well defined. Yet, $\Tb(x) = (\min(0,x), -\max(0,x))$ is differentiable almost everywhere, but $T_1'(\RR_{+}) = 0$ and $T_2'(\RR_{-})=0$. It follows that $p$ is not strongly exponential. 
\end{example}

\section{PROOFS}

\label{app:proofs}

\subsection{Proof of Definition~\ref{def:sim}}
\label{proof:sim}

\begin{proposition}
\label{prop:sim}
    The binary relations $\sim_A$ and $\sim_P$ are equivalence relations on $\Theta$.
\end{proposition}

The following proof applies to both $\sim_A$ and $\sim_P$ which we will simply denote by $\sim$.

It is clear that $\sim$ is reflexive and symmetric. 
Let $((\fb, \Tb, \lambdab), (\tilde{\fb}, \tilde{\Tb}, \tilde{\lambdab}), (\overline{\fb}, \overline{\Tb}, \overline{\lambdab})) \in \Theta^3$, s.t. $(\fb, \Tb, \lambdab) \sim (\tilde{\fb}, \tilde{\Tb}, \tilde{\lambdab})$ and $(\fb, \Tb, \lambdab) \sim (\overline{\fb}, \overline{\Tb}, \overline{\lambdab})$. Then $\exists A_1, A_2$ and $\mathbf{c}_1, \mathbf{c}_2$ s.t.
\begin{equation}
    \begin{aligned}
        \Tb(\fb^{-1}(\xb)) &= A_1 \tilde{\Tb}(\tilde{\fb}^{-1}(\xb)) + \mathbf{c}_1\textrm{ and} \\
        \overline{\Tb}(\overline{\fb}^{-1} (\xb)) &= A_2 \Tb(\fb^{-1}(\xb)) + \mathbf{c}_2\\
                        &= A_2A_1 \tilde{\Tb}(\tilde{\fb}^{-1}(\xb)) + A_2\mathbf{c}_1 + \mathbf{c}_2\\
                        &= A_3 \tilde{\Tb}(\tilde{\fb}^{-1}(\xb)) + \mathbf{c}_3
    \end{aligned}
\end{equation}
and thus $(\tilde{\fb}, \tilde{\Tb}, \tilde{\lambdab}) \sim (\overline{\fb}, \overline{\Tb}, \overline{\lambdab})$. \QED

\subsection{Proof of Theorem~\ref{th:iden}}

\subsubsection{Main steps of the proof}
The proof of this Theorem is done in three steps. 

In the first step, we use a simple convolutional trick made possible by assumption \ref{th:iden:ass1}, to transform the equality of observed data distributions into equality of noiseless distributions. In other words, it simplifies the noisy case into a noiseless case. This step results in equation $\eqref{eq:php}$.

The second step consists of removing all terms that are either a function of observations $\xb$ or auxiliary variables $\ub$. This is done by introducing the points provided by assumption \ref{th:iden:ass4}, and using $\ub_0$ as a "pivot". This is simply done in equations \eqref{eq:php}-\eqref{eq:php4}. 

The last step of the proof is slightly technical. The the goal is to show that the linear transformation is invertible thus resulting in an equivalence relation. This is where we use assumption \ref{th:iden:ass3}.  

\subsubsection{Proof}
\paragraph{Step I} We introduce here the volume of a matrix denoted $\vol A$ as the product of the singular values of $A$. When $A$ is full column rank, $\vol A = \sqrt{\det A^TA}$, and when $A$ is invertible, $\vol A = \snorm{\det A}$. The matrix volume can be used in the change of variable formula as a replacement for the absolute determinant of the Jacobian \citep{ben-israel1999changeofvariables}. This is most useful when the Jacobian is a rectangular matrix ($n < d$).
Suppose we have two sets of parameters $(\fb, \Tb, \lambdab)$ and $(\tilde{\fb}, \tilde{\Tb}, \tilde{\lambdab})$ such that $p_{\fb, \Tb, \lambdab}(\xb \vert \ub) = p_{\tilde{\fb}, \tilde{\Tb}, \tilde{\lambdab}}(\xb \vert \ub)$ for all pairs $(\xb, \ub)$.
Then:

\begin{align}
\label{eq:conv1}
    && \int_\Zcal p_{\Tb, \lambdab}(\zb\vert\ub) p_\fb(\xb\vert\zb) \dd\zb &= \int_\Zcal p_{\tilde{\Tb}, \tilde{\lambdab}}(\zb\vert\ub) p_{\tilde{\fb}}(\xb\vert\zb) \dd\zb \\
\label{eq:conv2}
    \Rightarrow && \int_\Zcal p_{\Tb, \lambdab}(\zb\vert\ub) p_\eps(\xb - \fb(\zb)) \dd \zb &= \int_\Zcal p_{\tilde{\Tb}, \tilde{\lambdab}}(\zb\vert\ub) p_\eps(\xb - \tilde{\fb}(\zb)) \dd \zb \\
\label{eq:conv3}
    \Rightarrow && \int_\XX p_{\Tb, \lambdab}(\fb^{-1}(\xbar)\vert\ub) \vol J_{\fb^{-1}}(\xbar) p_\eps(\xb - \xbar) \dd \xbar &= \int_\XX  p_{\tilde{\Tb}, \tilde{\lambdab}}(\tilde{\fb}^{-1}(\xbar)\vert\ub) \vol J_{\tilde{\fb}^{-1}}(\xbar) p_\eps(\xb - \xbar) \dd \xbar \\
\label{eq:conv4}
    \Rightarrow && \int_{\RR^d} \tilde{p}_{\Tb, \lambdab, \fb, \ub}(\xbar) p_\eps(\xb - \xbar) \dd\xbar &= \int_{\RR^d} \tilde{p}_{\tilde{\Tb}, \tilde{\lambdab}, \tilde{\fb}, \ub}(\xbar) p_\eps(\xb - \xbar) \dd \xbar \\
\label{eq:conv5}
    \Rightarrow && (\tilde{p}_{\Tb, \lambdab, \fb, \ub}*p_\eps)(\xb) &= (\tilde{p}_{\tilde{\Tb}, \tilde{\lambdab}, \tilde{\fb}, \ub}*p_\eps)(\xb) \\
\label{eq:conv6}
    \Rightarrow && F[\tilde{p}_{\Tb, \lambdab, \fb, \ub}](\omega) \varphi_\eps(\omega) &= F[\tilde{p}_{\tilde{\Tb}, \tilde{\lambdab}, \tilde{\fb}, \ub}](\omega) \varphi_\eps(\omega) \\
\label{eq:conv7}
    \Rightarrow && F[\tilde{p}_{\Tb, \lambdab, \fb, \ub}](\omega) &= F[\tilde{p}_{\tilde{\Tb}, \tilde{\lambdab}, \tilde{\fb}, \ub}](\omega) \\
\label{eq:conv8}
    \Rightarrow && \tilde{p}_{\Tb, \lambdab, \fb, \ub}(\xb) &= \tilde{p}_{\tilde{\Tb}, \tilde{\lambdab}, \tilde{\fb}, \ub}(\xb)
\end{align}

where:
\begin{itemize}
    \item in equation \eqref{eq:conv3}, $J$ denotes the Jacobian, and we made the change of variable $\xbar = \fb(\zb)$ on the left hand side, and $\xbar = \tilde{\fb}(\zb)$ on the right hand side.
    \item in equation \eqref{eq:conv4}, we introduced 
    \begin{equation}
        \tilde{p}_{\Tb, \lambdab, \fb, \ub}(\xb) = p_{\Tb, \lambdab}(\fb^{-1}(\xb)\vert\ub)\vol J_{\fb^{-1}}(\xb) \mathds{1}_{\XX}(\xb)
    \end{equation}
     on the left hand side, and similarly on the right hand side.
    \item in equation \eqref{eq:conv5}, we used $*$ for the convolution operator.
    \item in equation \eqref{eq:conv6}, we used $F[.]$ to designate the Fourier transform, and where $\varphi_\eps = F[p_\eps]$ (by definition of the characteristic function).
    \item in equation \eqref{eq:conv7}, we dropped $\varphi_\eps(\omega)$ from both sides as it is non-zero almost everywhere (by assumption \ref{th:iden:ass1}).
\end{itemize}
Equation \eqref{eq:conv8} is valid for all $(\xb, \ub) \in \XX\times\mathcal{U}$. What is basically says is that for the distributions to be the same after adding the noise, the noise-free distributions have to be the same. Note that $\xb$ here is a general variable and we are actually dealing with the noise-free probability densities.

\paragraph{Step II} By taking the logarithm on both sides of equation \eqref{eq:conv8} and replacing $p_{\Tb, \lambdab}$ by its expression from \eqref{eq:zcu}, we get:

\begin{multline}
\label{eq:php}
        \log \vol J_{\fb^{-1}}(\xb) + \sum_{i=1}^n ( \log Q_i(f_i^{-1}(\xb)) - \log Z_i(\ub)  +
        \sum_{j=1}^k T_{i,j}(f_i^{-1}(\xb)) \lambda_{i,j}(\ub)) = \\
        \log \vol J_{\tilde{\fb}^{-1}}(\xb) + \sum_{i=1}^n ( \log \tilde{Q}_i({\tilde{f}_i}^{-1}(\xb)) - \log \tilde{Z}_i(\ub)  +
        \sum_{j=1}^k \tilde{T}_{i,j}({\tilde{f}_i}^{-1}(\xb)) \tilde{\lambda}_{i,j}(\ub))
\end{multline}

Let $\ub_0, \dots, \ub_{nk}$ be the points provided by assumption \ref{th:iden:ass4} of the Theorem, and define $\overline{\lambdab}(\ub) = \lambdab(\ub) - \lambdab(\ub_0)$. We plug each of those $\ub_l$ in \eqref{eq:php} to obtain $nk+1$ such equations. We subtract the first equation for $\ub_0$ from the remaining $nk$ equations to get for $l = 1, \dots, nk$:
\begin{equation}
\label{eq:php2}
        \dotprod{\Tb(\fb^{-1}(\xb))}{\overline{\lambdab}(\ub_l)} + \sum_i \log \frac{Z_i(\ub_0)}{Z_i(\ub_l)} = \\ 
        \dotprod{\tilde{\Tb}(\tilde{\fb}^{-1}(\xb))}{\overline{\tilde{\lambdab}}(\ub_l)} + \sum_i \log \frac{\tilde{Z}_i(\ub_0)}{\tilde{Z}_i(\ub_l)} 
\end{equation}

Let $L$ bet the matrix defined in assumption \ref{th:iden:ass4}, and $\tilde{L}$ similarly defined for $\tilde{\lambdab}$ ($\tilde{L}$ is not necessarily invertible). Define $b_l = \sum_i \log \frac{\tilde{Z}_i(\ub_0)Z_i(\ub_l)}{Z_i(\ub_0)\tilde{Z}_i(\ub_l)}$ and $\mathbf{b}$ the vector of all $b_l$ for $l=1,\dots,nk$.  Expressing \eqref{eq:php2} for all points $\ub_l$ in matrix form, we get:
\begin{equation}
\label{eq:php3}
	L^T \Tb(\fb^{-1} (\xb)) = \tilde{L}^T \tilde{\Tb}(\tilde{\fb}^{-1}(\xb)) + \mathbf{b}
\end{equation}

We multiply both sides of \eqref{eq:php3} by the transpose of the inverse of $L^T$ from the left to find:
\begin{equation}
\label{eq:php4}
	\Tb(\fb^{-1}(\xb)) = A \tilde{\Tb}(\tilde{\fb}^{-1}(\xb)) + \mathbf{c}
\end{equation}
where $A = L^{-T}\tilde{L}$ and $\mathbf{c} = L^{-T} \mathbf{b}$.

\paragraph{Step III} Now by definition of $\Tb$ and according to assumption \ref{th:iden:ass3}, its Jacobian exists and is an $nk\times n$ matrix of rank $n$. This implies that the Jacobian of $\tilde{\Tb}\circ\tilde{\fb}^{-1}$ exists and is of rank $n$ and so is $A$. We distinguish two cases:

\begin{itemize}
    \item If $k=1$, then this means that $A$ is invertible (because $A$ is $n \times n$).
    \item If $k>1$, define $\xbar = \fb^{-1}(\xb)$ and $\Tb_i(\bar{x}_i) = (T_{i,1}(\bar{x}_i), \dots T_{i,k}(\bar{x}_i))$. According to Lemma \ref{lem:expfam}, for each $i \in [1, \dots, n]$ there exist $k$ points $\bar{x}^1_i, \dots, \bar{x}^k_i$ such that $(\Tb_i'(\bar{x}_i^1), \dots, \Tb_i'(\bar{x}_i^k))$ are linearly independent. Collect those points into $k$ vectors $(\xbar^1, \dots, \xbar^k)$, and concatenate the $k$ Jacobians $J_{\Tb}(\xbar^l)$ evaluated at each of those vectors horizontally into the matrix $Q = (J_{\Tb}(\xbar^1), \dots, J_{\Tb}(\xbar^k))$ (and similarly define $\tilde{Q}$ as the concatenation of the Jacobians of $\tilde{\Tb}(\tilde{\fb}^{-1}\circ\fb(\xbar))$ evaluated at those points). Then the matrix $Q$ is invertible (through a combination of Lemma \ref{lem:expfam} and the fact that each component of $\tilde{T}$ is univariate). By differentiating $\eqref{eq:php4}$ for each $\xb^l$, we get (in matrix form):
    \begin{equation}
        Q = A \tilde{Q}
    \end{equation}
    The invertibility of $Q$ implies the invertibility of $A$ and $\tilde{Q}$. 
\end{itemize}
 
Hence, \eqref{eq:php4} and the invertibility of $A$ mean that $(\tilde{\fb}, \tilde{\Tb}, \tilde{\lambdab}) \sim (\fb, \Tb, \lambdab)$.

Moreover, we have the following observations:
\begin{itemize}
    \item the invertibility of $A$ and $L$ imply that $\tilde{L}$ is invertible, 
    \item because the Jacobian of $\tilde{\Tb}\circ\tilde{\fb}^{-1}$ is full rank and $\tilde{\fb}$ is injective (hence its Jacobian is full rank too), $J_{\tilde{\Tb}}$ has to be full rank too, and $\tilde{T}_{i,j}'(z) \neq 0$ almost everywhere.
    \item the real equivalence class of identifiability may actually be narrower that what is defined by $\sim$, as the matrix $A$ and the vector $\mathbf{c}$ here have very specific forms, and are functions of $\lambdab$ and $\tilde{\lambdab}$. \QED
\end{itemize}

\subsubsection{Understanding assumption \ref{th:iden:ass4} in Theorem \ref{th:iden}}
\label{app:understanding}

Let $\ub^0$ be an arbitrary point in its support $\mathcal{U}$, and $h(\ub) = \left( \lambda_{1,1}(\ub) - \lambda_{1,1}(\ub^0), \dots, \lambda_{n,k}(\ub) - \lambda_{n,k}(\ub^0) \right) \in \RR^{nk}$. Saying that there exists $nk$ distinct points $\ub^1$ to $\ub^{nk}$ (all different from $\ub^0$) such that $L$ is invertible is equivalent to saying that the vectors $\hb := (h(\ub^1), \dots, h(\ub^{nk}))$ are linearly independent in $\RR^{nk}$.
Let's suppose for a second that for any such choice of points, these vectors are not linearly independent. This means that $h(\mathcal{U})$ is necessarily included in a subspace of $\RR^{nk}$ of dimension at most $nk -1$. Such a subspace has measure zero in $\RR^{nk}$. Thus, if $h(\mathcal{U})$ isn't included in a subset of measure zero in $\RR^{nk}$, this can't be true, and there exists a set of points $\ub^1$ to $\ub^{nk}$ (all different from $\ub^0$) such that $L$ is invertible. This implies that as long as the $\lambda_{i,j}(\ub)$ are generated randomly and independently, then almost surely, $h(\mathcal{U})$ won't be included in any such subset with measure zero, and the assumption holds. 

We next gvive a simple example where this assumption always holds. Suppose $n=2$ and $k=1$, and that the auxiliary variable is a positive scalar. Consider sources $z_i \sim \NN(0, \lambda_i(u))$ that are distributed according to Gaussian distributions with zero mean and variances modulated as follows:
\begin{align}
    \lambda_1(u) &= u \\
    \lambda_2(u) &= u^2
\end{align}
Because the functions $u\mapsto u$ and $u\mapsto u^2$ are linearly independent (as functions), then for any choice of "pivot" point $u_0$, for instance $u_0 = 1$, and any choice of distinct non-zero scalars $u_1$ and $u_2$, the columns of the matrix $L:=(\lambdab(u_1) - 1, \lambdab(u_2) - 1)$ are linearly independent, and the matrix is invertible. 

\subsection{Proof of Theorem~\ref{th:iden2}}

\subsubsection{Main steps of the proof}
The proof of this Theorem is done in two main steps.

The first step is to show that $\tilde{\fb}^{-1}\circ\fb$ is a pointwise function. This is done by showing that the product of any two distinct partial derivatives of any component is always zero. Along with invertibility, this means that each component depends exactly on one variable. This is where we use the two additional assumptions required by the Theorem.

In the second step, we plug the result of the first step in the equation that resulted from Theorem \ref{th:iden} (see equation \eqref{eq:pth26}). The fact that $\Tb$, $\tilde{\Tb}$ and $\tilde{\fb}^{-1}\circ\fb$ are all pointwise functions implies that $A$ is necessarily a permutation matrix. 

\subsubsection{Proof}
\paragraph{Step I} In this Theorem we suppose that $k\geq 2$. The assumptions of Theorem~\ref{th:iden} hold, and so we have 
\begin{equation}
\label{eq:pth21}
	\Tb(\fb^{-1}(\xb)) = A \tilde{\Tb}(\tilde{\fb}^{-1}(\xb)) + \mathbf{c}
\end{equation}
for an invertible $A\in\RR^{nk\times nk}$. We will index $A$ by four indices $(i,l,a,b)$, where $1\leq i\leq n, 1\leq l \leq k$ refer to the rows and $1\leq a\leq n, 1\leq b \leq k$ to the columns. Let $\vb(\zb) = \tilde{\fb}^{-1}\circ\fb(\zb):\ZZ\rightarrow\ZZ$. Note that $\vb$ is bijective because $\fb$ and $\tilde{\fb}$ are injective. Our goal is to show that $v_i(\zb)$ is a function of only one $z_{j_i}$, for all $i$. We will denote by $v_i^s:=\pderiv{v_i}{z_s}(\zb)$, and $v_i^{st}:=\frac{\partial^2v_i}{\partial z_s\partial z_t}(\zb)$. For each $1\leq i\leq n$ and $1\leq l \leq k$, we get by differentiating \eqref{eq:pth21} with respect to $z_s$:
\begin{equation}
\label{eq:pth22}
    \delta_{is}T_{i,l}'(z_i) = \sum_{a,b} A_{i,l,a,b} \tilde{T}_{a,b}'(v_a(\zb))v_a^s(\zb)
\end{equation}
and by differentiating \eqref{eq:pth22} with respect to $z_t, t>s$: 
\begin{equation}
\label{eq:pth23}
    0 = \sum_{a,b} A_{i,l,a,b} \left( \tilde{T}_{a,b}'(v_a(\zb))v_a^{s,t}(\zb) \right. \\ 
    + \left. \tilde{T}_{a,b}''(v_a(\zb))v_a^s(\zb)v_a^t(\zb) \right)
\end{equation}
This equation is valid for all pairs $(s,t), t>s$. Define 
$\mathbf{B}_a(\zb):=\left(v_a^{1,2}(\zb), \dots, v_a^{n-1,n}(\zb)\right)\in\RR^{\frac{n(n-1)}{2}}$, 
$\mathbf{C}_a(\zb):=\left(v_a^{1}(\zb)v_a^{2}(\zb), \dots, v_a^{n-1}(\zb)v_a^{n}(\zb)\right)\in\RR^{\frac{n(n-1)}{2}}$, 
$M(\zb) := \left( \mathbf{B}_1(\zb), \mathbf{C}_1(\zb), \dots, \mathbf{B}_n(\zb), \mathbf{C}_n(\zb) \right)$, 
$\eb^{(a,b)} := (0, \dots, 0, T_{a,b}', T_{a,b}'', 0, \dots, 0)\in\RR^{2n}$, such that the non-zero entries are at indices $(2a, 2a+1)$ and 
$\overline{e}(\zb):=(\eb^{(1,1)}(z_1), \dots, \eb^{(1,k)}(z_1), \dots, \eb^{(n,1)}(z_n), \dots, \eb^{(n,k)}(z_n))$ $\in \RR^{2n \times nk}$. 
Finally, denote by $A_{i,l}$ the $(i,l)$-th row of $A$. Then by grouping equation~\eqref{eq:pth23} for all valid pairs $(s,t)$ and pairs $(i,l)$ and writing it in matrix form, we get:

\begin{equation}
\label{eq:pth24}
    M(\zb) \overline{e}(\zb)A = 0
\end{equation}

Now by Lemma \ref{lem:var_exp_fam}, we know that $\overline{e}(\zb)$ has rank $2n$ almost surely on $\ZZ$. Since A is invertible, it is full rank, and thus $\operatorname*{rank}(\overline{e}\left(\zb)A\right) = 2n$ almost surely on $\ZZ$. It suffices then to multiply by its pseudo-inverse from the right to get 
\begin{equation}
\label{eq:pth25}
    M(\zb) = 0
\end{equation}
In particular, $C_a(\zb)=0$ for all $1\leq a \leq n$. This means that the Jacobian of $\vb$ at each $\zb$ has at most one non-zero entry in each row. By invertibility and continuity of $J_\vb$, we deduce that the location of the non-zero entries are fixed and do not change as a function of $\zb$. This proves that $\tilde{\fb}^{-1}\circ\fb$ is point-wise nonlinearity. 

\paragraph{Step II} Let $\overline{T}(\zb) = \tilde{T}(\vb(\zb)) + A^{-1}\mathbf{c}$. $\overline{\Tb}$ is a composition of a permutation and pointwise nonlinearity. Without any loss of generality, we assume that the permutation in $\overline{T}$ is the identity. Plugging this back into equation \eqref{eq:pth21} yields:
\begin{equation}
\label{eq:pth26}
    \Tb(\zb) = A\overline{\Tb}(\zb)
\end{equation}
Let $D = A^{-1}$. The last equation is valid for every component:
\begin{equation}
\label{eq:pth27}
    \overline{T}_{i,l}(z_i) = \sum_{a,b}D_{i,l,a,b} T_{a,b}(z_a)
\end{equation}
By differentiating both sides with respect to $z_s$ where $s\neq i$ we get
\begin{equation}
\label{eq:pth28}
    0 = \sum_{b} D_{i,l,s,b}T_{s,b}'(z_s)
\end{equation}
By Lemma \ref{lem:suffstat_lin2}, we get $D_{i,l,s,b} = 0$ for all $1\leq b \leq k$. Since \eqref{eq:pth28} is valid for all $l$ and all $s \neq i$, we deduce that the matrix $D$ has a block diagonal form:
\begin{equation}
    D = \begin{pmatrix}
        D_1 &  &  \\
         & \ddots &  \\
         &  & D_n
    \end{pmatrix}
\end{equation}
We conclude that $A$ has the same block diagonal form. Each block $i$ transforms $\Tb_i(\zb)$ into $\overline{\Tb}_i(\zb)$, which achieves the proof.
\QED

\subsection{Proof of Theorem~\ref{th:iden3}}

\subsubsection{Main steps of the proof}
This proof uses concepts borrowed from differential geometry. A good reference is the monograph by \cite{lee2003introduction}. 

By defining $\vb = \fb^{-1}\circ\tilde{\fb}$, equation \eqref{eq:php4} implies that each function $T_i\circ v_i$ can be written as a separable sum, \ie a sum of $n$ maps where each map $h_{i,a}$ is function of only one component $z_a$. 

Intuitively, since $T_i$ is not monotonic, it admits a local extremum (supposed to be a minimum). By working locally around this minimum, we can suppose that it is global and attained at a unique point $y_i$. The smoothness condition on $\vb$ imply that the manifold where $T_i\circ v_i$ is minimized has dimension $n-1$. This is where we need assumption \ref{th:iden3:ass4} of the Theorem.

On the other hand, because of the separability in the sum, each non constant $h_{i,k}$ (minimized as a consequence of minimizing $T_i \circ v_i$) introduces a constraint on this manifold that reduces its dimension by $1$. That's why we can only have one non constant $h_{i,k}$ for each $i$. 

\subsubsection{Proof}
In this Theorem we suppose that $k=1$. For simplicity, we drop the exponential family component index: $T_i:=T_{i,1}$. By introducing $\vb = \fb^{-1}\circ\tilde{\fb}$ and $h_{i,a}(z_a) = A_{i,a}\tilde{T}_a(z_a) + \frac{c_i}{n}$ into equation \eqref{eq:php4}, we can rewrite it as:
\begin{equation}
    \label{eq:pth31}
    T_i(v_i(\zb)) = \sum_{a=1}^n h_{i,a}(z_a)
\end{equation}
for all $1\leq i \leq n$. 

By assumption, $h_{i,a}$ is not monotonic, and so is $T_i$. So for each $a$, there exists $\tilde{y}_{i,a}$ where $h_{i,a}$ reaches an extremum, which we suppose is a minimum without loss of generality.
This implies that $T_i\circ v_i$ reaches a minimum at $\tilde{\yb}_i:=(\tilde{y}_{i,1}, \dots, \tilde{y}_{i,n})$, which in turn implies that $y_i := v_i(\tilde{\yb}_i)$ is a point where $T_i$ reaches a local minimum. Let $U$ be an open set centered around $y_i$, and let $\tilde{V}:=v_i^{-1}[U]$ the preimage of $U$ by $v_i$. Because $v_i$ is continuous, $\tilde{V}$ is open in $\RR^n$ and non-empty because $\tilde{\yb}_i \in \tilde{V}$. We can then restrict ourselves to a cube $V \subset \tilde{V}$ that contains $\tilde{\yb}_i$ which can be written as $V=V_1\times \dots \times V_n$ where each $V_a$ is an open interval in $\RR$. 

We can chose $U$ such that $T_i$ has only one minimum that is reached at $y_i$. This is possible because $T_i' \neq 0$ almost everywhere by hypothesis. Similarly, we chose the cube $V$ such that each $h_{i,a}$ either has only one minimum that is reached at $\tilde{y}_{i,a}$, or is constant (possible by setting $A_{i,a}=0$). Define
\begin{align}
    m_{i} &= \min_{\zb\in V} T_i\circ v_i(\zb) \in \RR \\
    \mu_{i,a} &= \min_{z_a\in V_a} h_{i,a}(z_a) \in \RR
\end{align}
for which we have $m_i = \sum_a \mu_{i,a}$. 

Define the sets $C_{i} = \{ \zb \in V \vert T_i\circ v_i(\zb) = m_i \}$ , $\tilde{C}_{i,a} = \{ \zb \in V \vert h_{i,a}(z_a) = \mu_{i,a} \}$ and $\tilde{C}_{i} = \cap_a \tilde{C}_{i, a}$. We trivially have $\tilde{C}_{i} \subset C_{i}$. Next, we prove that $C_i \subset \tilde{C}_i$. Let $\zb \in C_{i}$, and suppose $\zb \notin \tilde{C}_{i}$. Then there exist an index $k$, $\eps \in \RR$ and $\tilde{\zb} = (z_1, \dots, z_k + \eps, \dots, z_n)$ such that $m_i = \sum_a h_{i,a}(z_a) > \sum_a h_{i,a}(\tilde{z}_a) \geq \sum_a \mu_{i,a} = m_i$ which is not possible. Thus $\zb \in \tilde{C}_{i}$. Hence, $\tilde{C}_{i} = C_{i}$. 

Since $m_i$ is only reached at $y_i$, we have $C_i = \{ \zb \in V \vert v_i(\zb) = y_i \}$. By hypothesis, $v_i$ is of class $\mathcal{C}^1$, and its Jacobian is non-zero everywhere on $V$ (by invertibility of $\vb$). Then, by Corollary 5.14 in \cite{lee2003introduction}, we conclude that $C_{i}$ is a smooth ($\mathcal{C}^1$) submanifold of co-dimension $1$ in $\RR^n$, and so is $\tilde{C}_i$ by equality. 

On the other hand, if $h_{i,a}$ is not constant, then it reaches its minimum $\mu_{i,a}$ at only one point $\tilde{y}_{i,a}$ in $V_a$. In this case, $\tilde{C}_{i,a} = V_{[1,i-1]}\times \{\tilde{y}_{i,a}\}\times V_{[i+1,n]}$. Suppose that there exist two different indices $a\neq b$, such that $h_{i,a}$ and $h_{i,b}$ are not constant. Then $\tilde{C}_{i,a}\cap \tilde{C}_{i,b}$ is a submanifold of co-dimension 2. This would contradict the fact that the co-dimension of $\tilde{C}_i$ is 1. 

Thus, exactly one of the $h_{i,a}$ is not constant for each $i$. This implies that the $i$-th row of matrix $A$ has exactly one non-zero entry. The non-zero entry should occupy a different position in each row to guarantee invertibility, which proves that $A$ is a scaled permutation matrix. Plugging this back into equation \eqref{eq:php4} implies that $\tilde{\fb}\circ\fb$ is a point-wise nonlinearity. \QED

\subsection{Proof of Proposition \ref{prop:nec}}
For simplicity, denote $Q(\zb) := \prod_iQ_i(z_i)$ and $Z(\ub) := \prod_iZ_i(\ub)$. Let $A$ be an orthogonal matrix and $\tilde{\zb} = A \zb$
It is easy to check that $\tilde{\zb}\sim p_{\tilde{\thetab}}(\tilde{\zb}\vert\ub)$ where this new exponential family is defined by the quantities $\tilde{Q} = Q$, $\tilde{\Tb}=\Tb$, $\tilde{\lambdab} = A\lambdab$ and $\tilde{Z} = Z$. In particular, the base measure $Q$ does not change when $Q_i(z_i)=1$ or $Q_i(z_i) = e^{-z_i^2}$ because such a $Q$ is a rotationally invariant function of $\zb$. Further, we have
\begin{equation}
    \dotprod{\zb}{\lambdab(\ub)} =\dotprod{A^T\tilde{\zb}}{\lambdab(\ub)} \\= \dotprod{\tilde{\zb}}{A\lambdab(\ub)} = \dotprod{\tilde{\zb}}{\tilde{\lambdab}(\ub)}
\end{equation}

Finally let $\tilde{\fb} = \fb\circ A^T$, and $\tilde{\thetab} := \left(\tilde{\fb}, \tilde{\Tb}, \tilde{\lambdab}\right)$. We get:
\begin{align}
    p_\thetab(\xb\vert\ub) &= \int p_\thetab(\xb\vert\zb) p_\thetab(\zb\vert\ub) \dd \zb \\
        &= \int p_\eps(x - \fb(\zb)) \frac{Q(\zb)}{Z(\ub)} \exp(\dotprod{\zb}{\lambdab(\ub)})\dd\zb\\
        \label{eq:proofnec2}
        &= \int p_\eps(x - \tilde{\fb}(\tilde{\zb})) \frac{\tilde{Q}(\tilde{\zb})}{\tilde{Z}(\ub)} \exp(\dotprod{\tilde{\zb}}{\tilde{\lambdab}(\ub)})\dd\tilde{\zb}\\
        &= p_{\tilde{\thetab}}(\xb\vert\ub)
\end{align}
where in equation \eqref{eq:proofnec2} we made the change of variable $\tilde{\zb} = A\zb$, and removed the Jacobian because it is equal to $1$. We then see that it is not possible to distinguish between $\thetab$ and $\tilde{\thetab}$ based on the observed data distribution. \QED

\subsection{Proof of Theorem \ref{th:vaeiden}}

The loss \eqref{eq:loss} can be written as follows:
\begin{equation}
    \LL(\thetab, \phib) = \log \pt(\xb\vert\ub) - \KL{\qp(\zb\vert\xb, \ub)}{\pt(\zb\vert\xb,\ub)}
\end{equation}
If the family $\qp(\zb\vert\xb, \ub)$ is large enough to include $\pt(\zb\vert\xb,\ub)$, then by optimizing the loss over its parameter $\phib$, we will minimize the KL term, eventually reaching zero, and the loss will be equal to the log-likelihood. The VAE in this case inherits all the properties of maximum likelihood estimation. In this particular case, since our identifiability is guaranteed up to equivalence classes, the consistency of MLE means that we converge to the equivalence class\footnote{this is easy to show: because true identifiability is one of the assumptions for MLE consistency, replacing it by identifiability up to equivalence class doesn't change the proof but only the conclusion.} (Theorem \ref{th:iden}) of true parameter $\thetab^*$ \ie in the limit of infinite data. \QED

%% file: sections/8_appendix_bis.tex
\section{DISCRETE OBSERVATIONS}
\label{app:discrete}

We can use a well-known logistic model to replace the additive Gaussian noise to model discrete observations. For example, in the binary case, let:
\begin{align}
    \mb &= \text{sigmoid}(\fb(\zb))\\
    \xb &\sim \text{Bernoulli}(\mb)
\end{align}
where $\text{sigmoid}()$ is the element-wise sigmoid nonlinearity.

However, by the very nature of discrete variables, the mapping $\zb \rightarrow \xb$ can no longer be injective. 
This is one of the key assumptions in our identifiability theory, which can no longer hold. 
The discrete observation case requires a bespoke identifiability proof. Nevertheless, we provide experiments in Supplementary Material~\ref{app:extra_exp} that strongly suggest that identifiability is achievable in such setting.
\section{UNIDENTIFIABILITY OF GENERATIVE MODELS WITH UNCONDITIONAL PRIOR}
\label{app:unident}

In this section, we present two well-known  proofs of unidentifiability of generative models. The first proof is simpler and considers factorial priors, which are widely-used in deep generative models and the VAE literature. The second proof is extremely general, and shows how any random vector can be transformed into independent components, in particular components which are standardized Gaussian. Thus, we see how in the general nonlinear case, there is little hope of finding the original latent variables based on the (unconditional, marginal) statistics of $\xb$ alone.

\subsection{Factorial priors}

Let us start with factorial, Gaussian priors. In other words, let $\zb \sim p_{\thetab}(\zb)=N(\mathbf{0},\mathbf{I})$. Now, a well-known result says that any orthogonal transformation of $\zb$ has exactly the same distribution. Thus, we could transform the latent variable by any orthogonal transformation $\zb'=M\zb$, and cancel that transformation in $p(\xb|\zb)$ (e.g.\ in the first layer of the neural network), and we would get exactly the same observed data (and thus obviously the same distribution of observed data) with $\zb'$.

\newcommand{\xib}{\boldsymbol{\xi}}
Formally we have
\begin{align}
    p_{\zb'}(\xib) &= p_{\zb}(M^T\xib)|\det M|=\frac{1}{(2\pi)^{d/2}}\exp(-\frac{1}{2}\|M^T\xib\|^2)\\
                    &=\frac{1}{(2\pi)^{d/2}}\exp(-\frac{1}{2}\|\xib\|^2)=p_{\zb}(\xib)
\end{align}
where we have used the fact that the determinant of an orthogonal matrix is equal to unity.

This result applies easily to any factorial prior. For $z_i$ of any distribution, we can transform it to a uniform distribution by $F_i(z_i)$ where $F_i$ is the cumulative distribution function of $z_i$. Next, we can transform it into standardized Gaussian by $\Phi^{-1}(F_i(z_i))$ where $\Phi$ is the standardized Gaussian cdf. After this transformation, we can again take any orthogonal transformation without changing the distribution. And we can even transform back to the same marginal distributions by $F_i^{-1}(\Phi(.))$. Thus, the original latents are not identifiable.

\subsection{General priors}

The second proof comes from the theory of nonlinear ICA \citep{hyvarinen1999nonlinear}, from which the following Theorem is adapted.

\begin{theorem}[\cite{hyvarinen1999nonlinear}]
\label{th:idenica}
Let $\zb$ be a $d$-dimensional random vector of any distribution. Then there exists a transformation  $\gb:\RR^d \rightarrow \RR^d$  such that the components of $\zb' := \gb(\zb)$  are independent, and each component has a standardized Gaussian distribution. In particular, $z_1'$ equals a monotonic transformation of $z_1$.
\end{theorem}
The proof is based on an iterative procedure reminiscent of Gram-Schmidt, where a new variable can always be transformed to be independent of any previously considered variables, which is  why $z_1$ is essentially unchanged.

This Theorem means that there are infinitely many ways of defining independent components $\zb$ that nonlinearly generated an observation $\xb$. This is because we can first transform $\zb$ any way we like and then apply the Theorem. The arbitrariness of the components is seen in the fact that we will always find that one arbitrary chosen variable in the transformation is one of the independent components. This is in some sense an alternative kind of indeterminacy to the one in the previous subsection.

In particular, we can even apply this Theorem on the observed data, taking $\xb$ instead of $\zb$. Then, in the case of factorial priors, just permuting the data variables, we would arrive at the conclusion that any of the $x_i$ can be taken to be one of the independent components, which is absurd. 

Now, to apply this theory in the case of a general prior on $\zb$, it is enough to point out that we can transform any variable into independent Gaussian variables, apply any orthogonal transformation, then invert the transformation in the Theorem, and we get a nonlinear transformation $\zb'=\gb^{-1}(M \gb(\zb))$ which has exactly the same distribution as $\zb$ but is a complex nonlinear transformation.  Thus, no matter what the prior may be, by looking at the data alone, it is not possible to recover the true latents based an unconditional prior distribution, in the general nonlinear case.  

\section{ALTERNATIVE FORMULATION OF THEOREM \ref{th:iden}}
\label{app:altiden}

\begin{theorem}
\label{th:altiden}

Assume that we observe data sampled from a generative model defined according to \eqref{eq:gen}-\eqref{eq:zcu}, with parameters $(\fb, \Tb, \lambdab)$. Assume the following holds:

\begin{enumerate}[label=(\roman*)]
    \item \label{th:altiden:ass1} The set $\{\xb \in \XX \vert \varphi_\eps (\xb) = 0\}$ has measure zero, where $\varphi_\eps$ is the characteristic function of the density $p_\eps$ defined in \eqref{eq:xcz}.
    \item \label{th:altiden:ass4} The mixing function $\fb$ in \eqref{eq:xcz} is injective.
    \item \label{th:altiden:ass2} The sufficient statistics $T_{i,j}$ in \eqref{eq:zcu} are differentiable almost everywhere, and $T_{i,j}' \neq 0$ almost everywhere for all $1 \leq i \leq n$ and $1 \leq j \leq k$.
	\item \label{th:altiden:ass3} $\lambdab$ is differentiable, and there exists $\ub_0 \in \mathcal{U}$ such that $J_\lambdab(\ub_0)$ is invertible.
\end{enumerate}
then the parameters $(\fb, \Tb, \lambdab)$ are $\sim$-identifiable.
Moreover, if there exists $(\tilde{\fb}, \tilde{\Tb}, \tilde{\lambdab})$ such that $p_{\tilde{\fb}, \tilde{\Tb}, \tilde{\lambdab}}(\xb \vert \ub) = p_{\fb, \Tb, \lambdab}(\xb \vert \ub)$, then $\tilde{\Tb}$ and $\tilde{\lambdab}$ verify assumptions $\ref{th:altiden:ass2}$ and $\ref{th:altiden:ass3}$.
\end{theorem}

\textit{Proof:}
The start of the proof is similar to the proof of Theorem \ref{th:iden}. When we get to equation \eqref{eq:php}:
\begin{multline}
        \log \vol J_{\fb^{-1}}(\xb) + \sum_{i=1}^n ( \log Q_i(f_i^{-1}(\xb)) - \log Z_i(\ub)  + \sum_{j=1}^k T_{i,j}(f_i^{-1}(\xb)) \lambda_{i,j}(\ub)) = \\
        \log \vol J_{\tilde{\fb}^{-1}}(\xb) + \sum_{i=1}^n ( \log \tilde{Q}_i({\tilde{f}_i}^{-1}(\xb)) - \log \tilde{Z}_i(\ub)  + \sum_{j=1}^k \tilde{T}_{i,j}({\tilde{f}_i}^{-1}(\xb)) \tilde{\lambda}_{i,j}(\ub))
\end{multline}
we take the derivative of both sides with respect to $\ub$ (assuming that $\tilde{\lambda}$ is also differentiable). All terms depending on $\xb$ only disappear, and we are left with:
\begin{equation}
    J_\lambdab(\ub)^T\Tb(\fb^{-1}(\xb)) - \sum_i\nabla \log Z_i(\ub) =  J_{\tilde{\lambdab}}(\ub)^T\tilde{\Tb}(\tilde{\fb}^{-1}(\xb)) - \sum_i\nabla \log \tilde{Z}_i(\ub)
\end{equation}
By evaluating both sides at $\ub_0$ provided by assumption \ref{th:altiden:ass3}, and multiplying both sides by $J_\lambdab(\ub_0)^{-T}$ (invertible by hypothesis), we find:
\begin{equation}
    \Tb(\fb^{-1}(\xb)) = A\tilde{\Tb}(\tilde{\fb}^{-1}(\xb)) + \mathbf{c}
\end{equation}
where $A = J_\lambdab(\ub_0)^{-T}J_{\tilde{\lambdab}}(\ub_0)^T$ and $\mathbf{c} = \sum_i\nabla \log \frac{Z_i(\ub_0)}{\tilde{Z}_i(\ub_0)}$.
The rest of the proof follows proof of Theorem \ref{th:iden}, where in the last part we deduce that $J_{\tilde{\lambdab}}(\ub_0)$ is invertible. \QED

\section{LINK BETWEEN MAXIMUM LIKELIHOOD AND TOTAL CORRELATION}
\label{app:mlmi}

Consider the noiseless case:
\begin{align}
    \xb &= \fb(\zb) \\
    p(\zb\vert\ub) &= \prod_i p_i(z_i\vert\ub)
\end{align}
where the components of the latent variable are independent given the auxiliary variable $\ub$. We can relate the log-likelihood of the data to the total correlation of the latent variables. To see this connection, let's use the change of variable formula in the expression of the log-likelihood:
\begin{align}
    \EE_{p(\xb, \ub)}\left[ \log p(\xb\vert\ub) \right] &=  \EE_{p(\zb, \ub)} \left[ \sum_i \log p_i(z_i\vert\ub) - \log \snorm{J_{\fb}(\zb)} \right] \\ 
\label{eq:lktc}
    &= - \EE_{p(\zb, \ub)} \left[\log \snorm{J_{\fb}(\zb)} \right] -\sum_i H(z_i\vert\ub) 
\end{align}
where $H(z_i\vert\ub)$ is the conditional differential entropy of $z_i$ given $\ub$. The same change of variable formula applied to $H(\xb\vert\ub)$ yields:
\begin{equation}
    H(\xb\vert\ub) = H(\zb\vert\ub) + \EE_{p(\zb, \ub)} \left[\log \snorm{J_{\fb}(\zb)} \right]
\end{equation}
which we then use in the expression of the conditional total correlation:
\begin{equation}
\label{eq:tc}
\begin{aligned}
    \TC(\zb\vert\ub) &:= \sum_i H(z_i\vert\ub) - H(\zb\vert\ub) \\
            &= \sum_i H(z_i\vert\ub) - H(\xb\vert\ub) + \EE_{p(\zb, \ub)} \left[\log \snorm{J_{\fb}(\zb)} \right]
\end{aligned}
\end{equation}
Putting equations \eqref{eq:lktc} and \eqref{eq:tc} together, we get:
\begin{equation}
    \EE_{p(\xb, \ub)}[\log p(\xb\vert\ub)] = - \TC(\zb\vert\ub) - H(\xb\vert\ub)
\end{equation}
The last term in this equation is a function of the data only and is thus a constant. An algorithm which learns to maximize the data likelihood is decreasing the total correlation of the latent variable. The total correlation is measure of independence as it is equal to zero if and only if the components of the latent variable are independent. 
Thus, by using a VAE to maximize a lower bound on the data likelihood, we are trying to learn an 
estimate of the inverse of the mixing function that gives the most independent components.

\section{REMARKS ON PREVIOUS WORK}
\label{app:prev}

\subsection{Previous work in nonlinear ICA}
\label{app:prev_ica}

\paragraph{ICA by Time Contrastive Learning} Time Contrastive Learning (TCL) introduced in \citet{hyvarinen2016unsupervised} is a method for nonlinear ICA based on the assumption that while the sources are independent, they are also non-stationary time series. This implies that they can be divided into
known non-overlapping segments, such that their distributions vary across segments. The non-stationarity is supposed to be slow compared to the sampling rate, so that we can consider the distributions within each segment to be unchanged over time; resulting in a 
piecewise stationary distribution across segments. Formally, given a segment index $\tau \in \mathcal{T}$, where $\mathcal{T}$ is a finite set of indices, the distribution of the sources within that segment is modelled as an exponential family, which is in their notation: 
\begin{equation}
\label{eq:tcl}
	\log p_\tau(\sbf) := \log p(\sbf \vert \mathrm{segment} = \tau) = \sum_{j=1}^d  \lambda_j(\tau) q(s_j) - \log Z(\tau)
\end{equation}
where $q_{j,0}$ is a stationary baseline and $q$ is the sufficient statistic for the exponential family of the sources (note that exponential families with different sufficient statistics for each source, or more than one sufficient statistic per source are allowed, but we focus on this simpler case here). Note that parameters $\lambda_j$ depend on the segment index, indicating that the distribution of sources changes across segments. It follows from equation \eqref{eq:mixing} that the observations are piece-wise stationary. 

TCL recovers the inverse transformation $\mathbf{f}^{-1}$ by self-supervised learning, where the goal is to classify original data points against segment indices in a multinomial classification task. To this end, TCL employed a deep network 
consisting of a feature extractor $h(\xb^{(i)}; \eta)$ with parameters $\eta$ in the form of a neural network, followed by a final classifying layer (e.g. softmax).
The theory of TCL, as stated in Theorem 1 of \cite{hyvarinen2016unsupervised}, is premised on the fact that in 
order to optimally classify observations into their corresponding 
segments the feature extractor, $h(\xb^{(i)}; \eta)$, 
must learn about the changes in the underlying distribution of 
latent sources. The theory shows that the method can learn the independent components up to transformations by sufficient statistics and a linear transformation, as in $\sim_A$ identifiability. It is further proposed that a linear ICA can recover the final $A$ if the number of segments grows infinite and the segment distributions are random in a certain sense, but this latter assumption is unrealistic in applications where the number of segments is small. We also emphasize that our estimation method based on VAE is very different from such a self-supervised scheme.

\paragraph{ICA using auxiliary variables}
A more recent development in nonlinear ICA is given by \cite{hyvarinen2019nonlinear} where it is assumed  that we observe data following a noiseless conditional nonlinear ICA case:
\begin{align}
    \xb &= \fb(\zb) \\
    p(\zb\vert\ub) &= \prod_i p_i(z_i\vert\ub)
\end{align}
This formulation is so general that it subsumes previous models by \cite{hyvarinen2016unsupervised, hyvarinen2017nonlinear} in the sense of the data model. However, their estimation method is very different from TCL: They rely on a self-supervised binary discrimination task based on randomization to learn the unmixing function. More specifically, from a dataset of observations and auxiliary variables pairs $\mathcal{D} = \{ \xb^{(i)}, \ub^{(i)}\}$, they construct a randomized dataset $\mathcal{D^{*}} = \{ \xb^{(i)}, \ub^{*}\}$ where $\ub^*$ is randomly drawn from the observed distribution of $\ub$. To distinguish between both datasets, a deep logistic regression is used. The last hidden layer of the neural network is a feature extractor denoted $\hb(\xb)$; like in TCL, the purpose of the feature extractor is therefore to extract the relevant features which will allow to distinguish between the two datasets. The identifiability results by \citet{hyvarinen2019nonlinear} have a lot of similarity to ours, and several of our proofs are inspired by them. However, we strengthen those results, while concentrating on the case of exponential family models. In particular, we show how any non-monotonic sufficient statistics for $k=1$ leads to identifiability in Theorem~\ref{th:iden3}, and also Theorem~\ref{th:iden2} generalizes the corresponding result (Theorem 2, case 2) in \citet{hyvarinen2019nonlinear}. Again, their estimation method is completely different from ours.

\subsection{Previous work on identifiability in VAEs}
Our framework might look similar to semi-supervised learning methods in the VAE context, due to the inclusion of the auxiliary variable $\ub$. However, the auxiliary variable $\ub$ can play a more general role. For instance, in time-series, it can simply be the the time index or history; in audiovisual data, it can be either one of the modalities, where the other is used as an observation. More importantly, and to our knowledge, there is no proof of identifiability in the semi-supervised literature. 

The question of identifiability, or lack of, in deep latent variable models especially VAEs has been tackled in work related to disentanglement. In \cite{mathieu2018disentangling,rolinek2018variational,locatello2018challenging} the authors show how isotropic priors lead to rotation invariance in the ELBO. We proved here (section~\ref{sec:identifiability} and supplementary material \ref{app:unident}) a much more general result: unconditional priors lead to unidentifiable models. These papers however focused on showcasing this problem, or how it can avoided in practice, and didn't provide alternative models that can be shown to be identifiable. This is what we try to achieve in this work, to provide a complementary analysis to previous research. Our proof of identifiability applies to the generative model itself, regardless of the estimation method. This is why we didn't focus in our analysis on the role of the encoder, which has been claimed to have a central role in some of the work cited above.

%% file: sections/9_appendix_figs.tex
\section{SIMULATION DETAILS}

\begin{figure*}[!ht]
\begin{subfigure}{.49\textwidth}
    \center{\includegraphics[width=\textwidth]
    {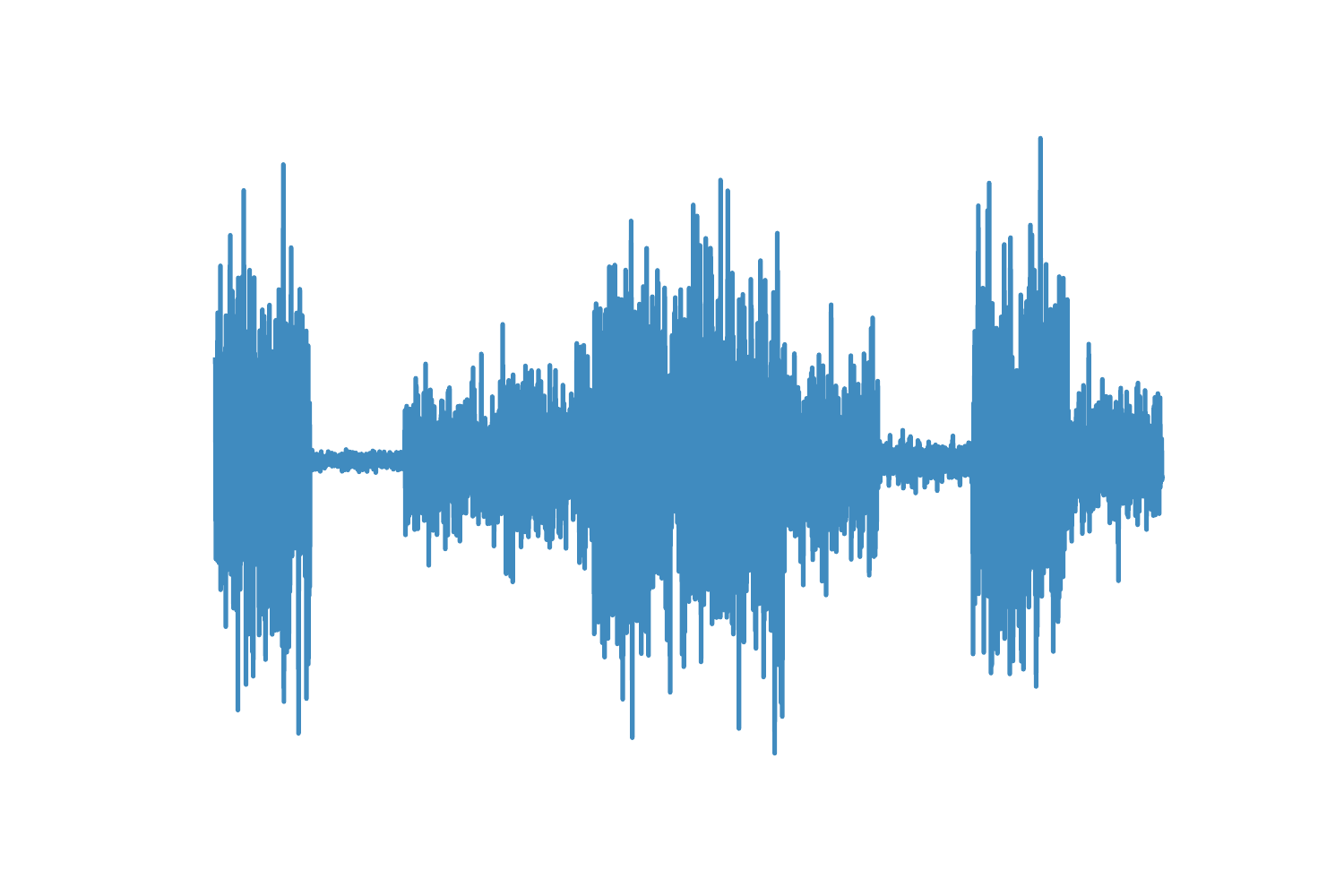}}
    \caption{\label{fig:example_normal}}
\end{subfigure}%
\begin{subfigure}{.49\textwidth}
    \center{\includegraphics[width=\textwidth]
    {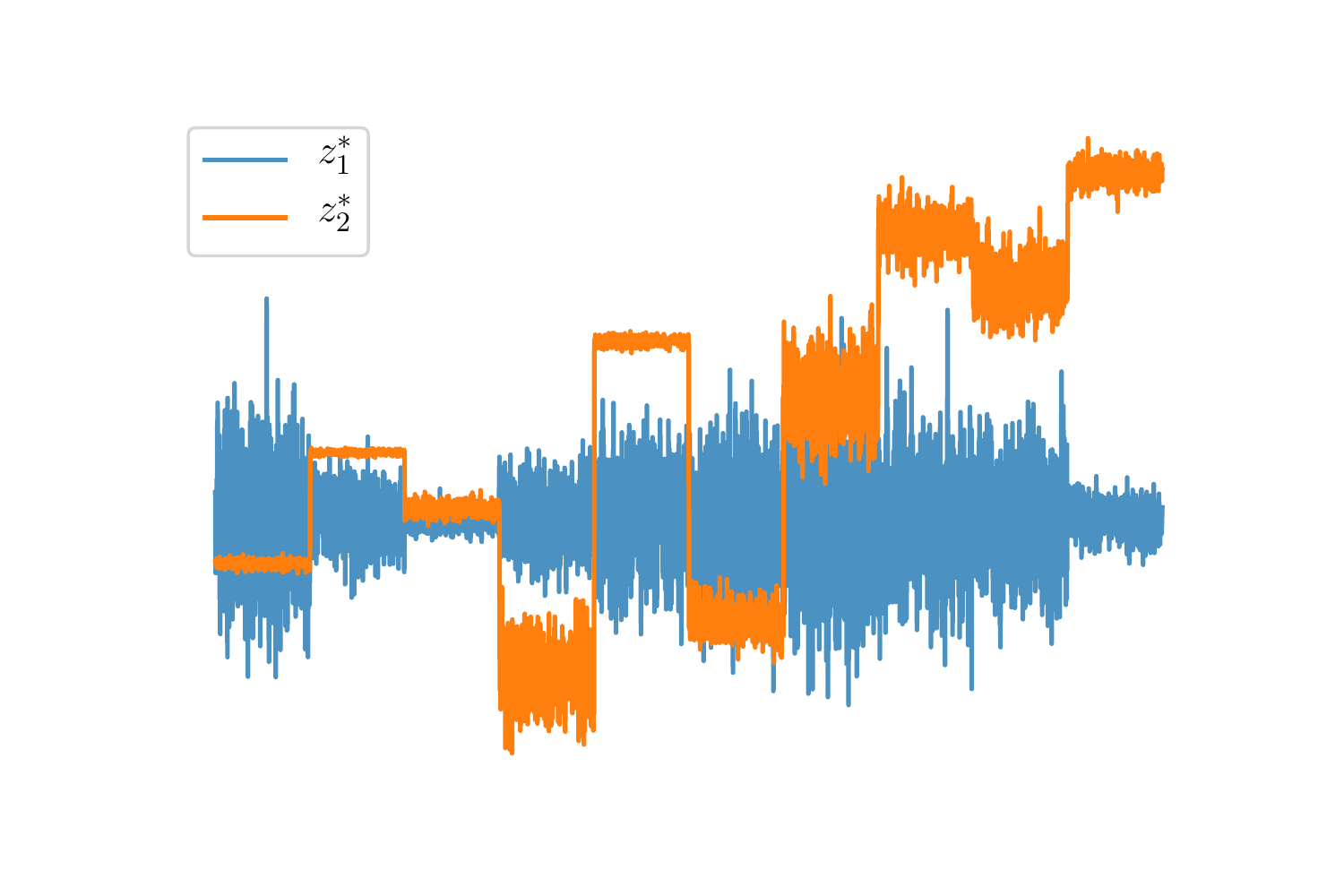}}
\caption{\label{fig:example_staircase}}
\end{subfigure}
\caption{\label{fig:examle_sources}
Visualization of various sources following the 
generative distribution detailed in equation (\ref{eq:zcu}).
$(a)$ single source with segment modulated variance; 
$(b)$ two sources where the mean of the
second source, $z_2^*$, is significantly modulated 
as a function of the segment, thus potentially serving to  greatly facilitate the 
surrogate classification task performed in TCL.  }
\end{figure*}

\subsection{Implementation detail for VAE experiments}
\label{app:model_details}

We give here more detail on the data generation process for our simulations. The dataset is described in section \ref{simulation}. 
The conditioning variable $\ub$ is the segment label, and its distribution is uniform on the integer set $\iset{1,M}$. Within each segment, the conditional prior distribution is chosen from the family \eqref{eq:zcu}, where $k=1$, $T_{i,1}(z_i) = z_i ^2$ and $Q_i(z_i) =1$, and the true $\lambda_i$ were randomly and independently generated across the segments and the components so that the variances have a uniform distribution on $[.5, 3]$. We sample latent variable $\zb$ from these distribution, and then mix them using a 4-layer multi-layer perceptron (MLP). An example of what the sources look like is plotted in Figure \ref{fig:example_normal}. We finally add small noise ($\sigma^2=0.01$) to the observations. When comparing to previous ICA methods, we omit this step, as these methods are for the noiseless case.

For the decoder \eqref{eq:xcz}, we chose $p_\eps = \NN\left(0, \sigma^2I \right)$ a zero mean Gaussian, where the scalar $\sigma^2$ controls the noise level. We fix the noise level $\sigma^2=0.01$. As for the inference model, we let $\qp(\zb\vert\xb,\ub) = \mathcal{N}\left(\zb|\gb(\xb, \ub; \phi_\gb), \diag{\sigmab^2(\xb, \ub; \phi_\sigmab)}\right)$ be a multivariate Gaussian with a diagonal covariance. The functional parameters of the decoder ($\fb$) and the inference model ($\gb$, $\sigmab^2$) as well as the conditional prior ($\lambdab$) are chosen to be MLPs, where the dimension of the hidden layers is varied between $10$ and $200$, the activation function is a leaky ReLU, and the number of layers is chosen from $\{3, 4, 5, 6 \}$. Mini-batches are of size 64, and the learning rate of the Adam optimizer is chosen from $\{0.01, 0.001\}$. We also use a scheduler to decay the learning rate as a function of epochs.

To implement the VAE, we followed \cite{kingma2013autoencoding}. We made sure the range of the hyperparameters (mainly number of layers and dimension of hidden layers) of the VAE is large enough for it to be comparable in complexity to our method (which has the extra $\lambda$ network to learn). 
To implement a $\beta$-VAE, we followed the instructions of \cite{higgins2016betavae} for the choice of hyperparameter $\beta$, which was chosen in the set $[1, 45]$.
Similarly, we followed \cite{chen2018isolating} for the choice of the hyperparameters $\alpha$, $\beta$ and $\gamma$ when implementing a $\beta$-TC-VAE: we chose $\alpha=\gamma=1$ and $\beta$ was chosen in the set $[1, 35]$.

\subsection{Description of significant mean modulated data}
\label{app:easy_data}
Here, we generated non-stationary $2$D data from a modified dataset as follows: $\zb^*\vert u\sim \NN(\mub(u), \diag(\sigmab^2(u))$ where $u$ is the segment index, $\mu_1(u) = 0$ for all $u$ and $\mu_2(u) = \alpha \gamma(u)$ where $\alpha\in\RR$ and $\gamma$ is a permutation. Essentially, the mean of the second source, $z_2^*$,
is significantly modulated by the segment index.
An example is plotted in Figure \ref{fig:example_staircase}. 
The variance $\sigmab^2(u)$ is generated randomly and independently across the segments. We then mix the sources into observations $\xb$ such that $x_1 = \textbf{MLP}(z_1, z_2)$ and $x_2 = z_2^*$, thus preserving the 
significant modulation of the mean in $x_2$. We note that this is just one of many potential mappings from $\mathbf{z}$ to $\mathbf{x}$ which could have been employed to yield significant mean modulation in $x_2$ across segments. 
TCL learns to unmix observations, $\mathbf{x}$, by solving a surrogate classification task. Formally, TCL seeks to 
train a deep network to accurately classify each observation into its corresponding segment. 
As such, the aforementioned dataset is designed to highlight the following 
limitation of TCL: due to its reliance on optimizing a 
self-supervised objective, it can fail to recover latent variables when the 
associated task is too easy. 
In fact, by choosing a large enough value of the separation parameter $\alpha$ (in our experiments $\alpha=2$), it is possible to classify samples by looking at the mean of $x_2$.

\section{FURTHER EXPERIMENTS}
\label{app:extra_exp}

\subsection{Additional general nonlinear ICA experiments}

As discussed in section \ref{ica}, our estimation method has many benefits over previously proposed self-supervised nonlinear ICA methods: it allows for dimensionality reduction, latent dimension selection based on the ELBO as a cross validation metric, and solving discrete ICA. We performed a series of simulations to test these claims.

\begin{figure}[!htb]
\begin{subfigure}{.32\textwidth}
    \center{\includegraphics[width=.94\textwidth]{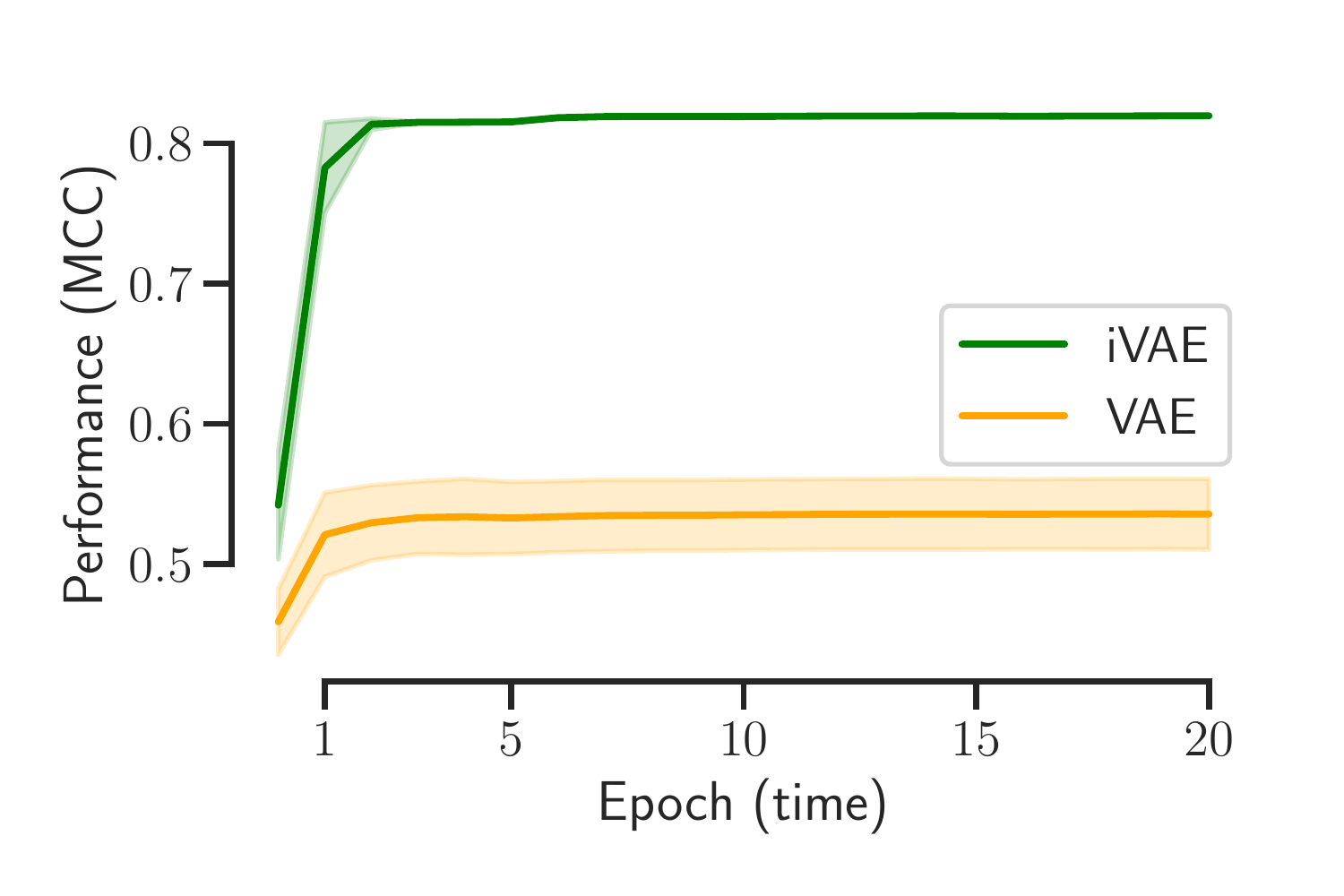}}
    \caption{\label{fig:ivae_discrete} iVAE for discrete ICA}
\end{subfigure}%
\begin{subfigure}{.32\textwidth}
  \center{\includegraphics[width=.94\textwidth]
    {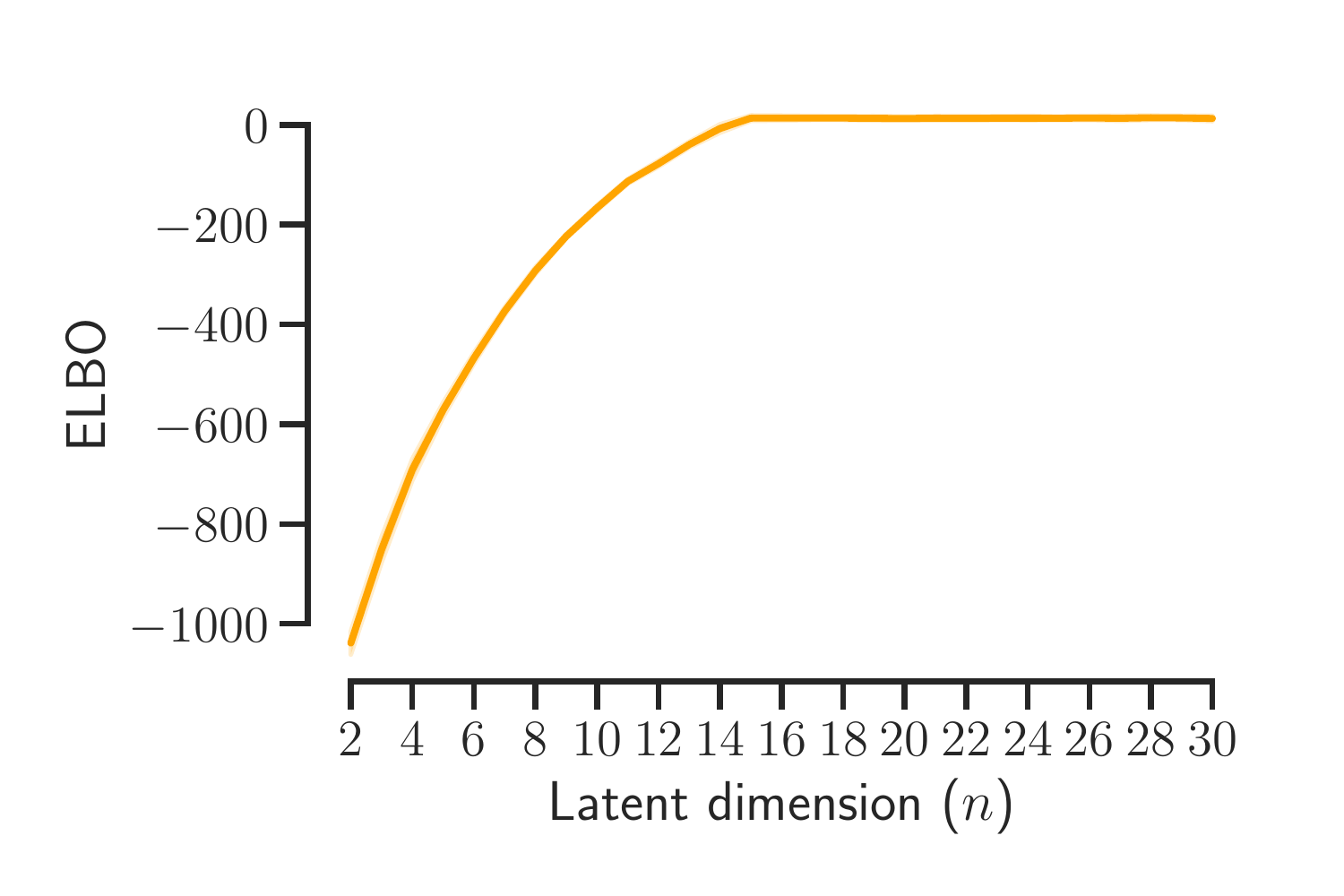}}
    \caption{\label{fig:elbo_dim_crossval_full} Dimensionality selection}
\end{subfigure}%
\begin{subfigure}{.32\textwidth}
    \center{\includegraphics[width=.94\textwidth]{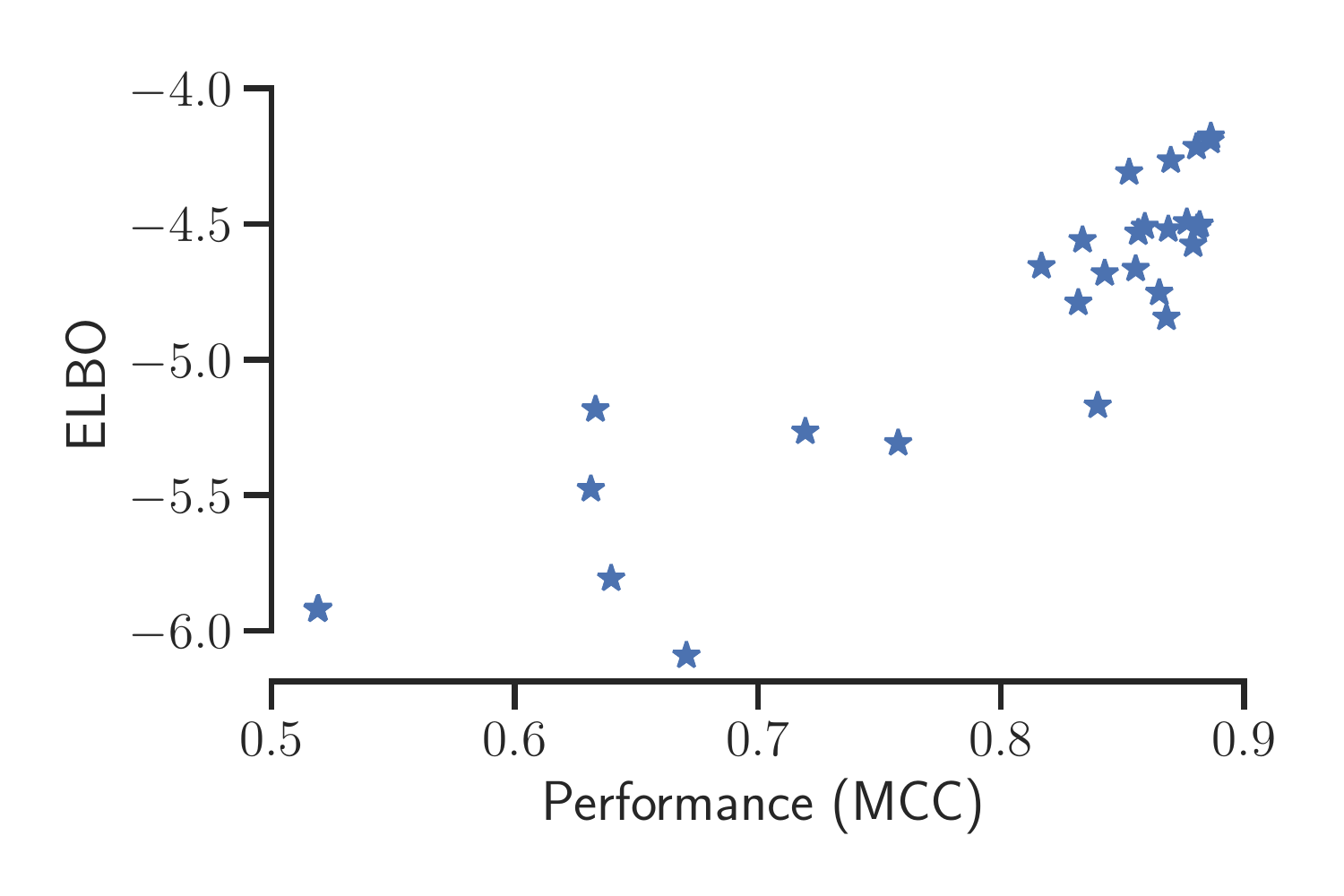}}
    \caption{\label{fig:elbo_crossval} Hyperparameter selection}
\end{subfigure}
\caption{\label{fig:ivae_good} $(a)$ Performance of iVAE and VAE on discrete ICA task. $(b)$ Evolution of the post training ELBO as a function of the latent dimension. The real dimension of the data is $d^*=80$ and the real dimension of the latent space is $n^*=15$. We observe an elbow at around $15$, thus successfully guessing the real dimension. $(c)$ ELBO as a function of the performance. Each star is an experiment run for a different set of hyperparameters.}
\end{figure}

\paragraph{Discrete observations}
To further test the capabilities of our method, we tested it on discrete data, and compared its identifiability performance to a vanilla VAE. The dimensions of the data and latents are $d=100$ and $n=10$.
The results are shown in Figure \ref{fig:ivae_discrete} and proves that our method is capable of performing discrete ICA.

\paragraph{Dimensionality selection and reduction}
The examples in section \ref{simulation} already showcased dimensionality reduction. In Figure \ref{fig:tcl_dim} for example, we have a mismatch between the dimensions of the latents and observations. In real world ICA applications, we usually don't know the dimension of the latents beforehand. One way to guess it is to use the ELBO as a proxy to select the dimension. Our method enables this when compared to previous nonlinear ICA methods like TCL \citep{hyvarinen2016unsupervised}. This is showcased in Figure \ref{fig:elbo_dim_crossval_full}, where the real dimensions of the simulated data are $d^*=80$ and $n^*=15$, and we run multiple experiments where we vary the latent dimensions between 2 and 40. We can see that the ELBO can be a good proxy for dimension selection, since it has a "knee" around the right value of dimension.

\paragraph{Hyperparameter selection} 
One important benefit of the proposed method is that it 
seeks to optimize an objective function derived from the marginal
log-likelihood of observations. As such, it follows that we may 
employ the ELBO to perform 
hyperparameter selection. 
To verify this claim, we run experiments for various distinct choices of hyperparameters (for example the dimension of hidden layers, number of hidden layers in the estimation network, learning rate, nonlinearities) on a synthetic dataset.
Results are provided in 
 Figure \ref{fig:elbo_crossval} which serves to 
empirically demonstrate that the ELBO is indeed a good proxy for how accurately we are able to recover the true latent variables.
In contrast, alternative methods for nonlinear ICA, such as TCL, 
do not provide principled and reliable proxies which reflect the accuracy of
estimated latent sources.

\subsection{Additional causality experiments for comparison to TCL}
\label{app:causal}

\paragraph{Setup}
The data generation process used by \cite[Section 4]{monti2019causal} is similar to the one we described in section \ref{simulation}, with the difference that the mixing should be in such a way that we get an acyclic causal relationship between the observations. This can be achieved 
by ensuring weight matrices in the mixing network are 
all lower-triangular, thereby introducing 
acyclic causal structure over observations. 

\begin{figure}
\begin{subfigure}{.32\textwidth}
    \center{\includegraphics[width=\textwidth]
    {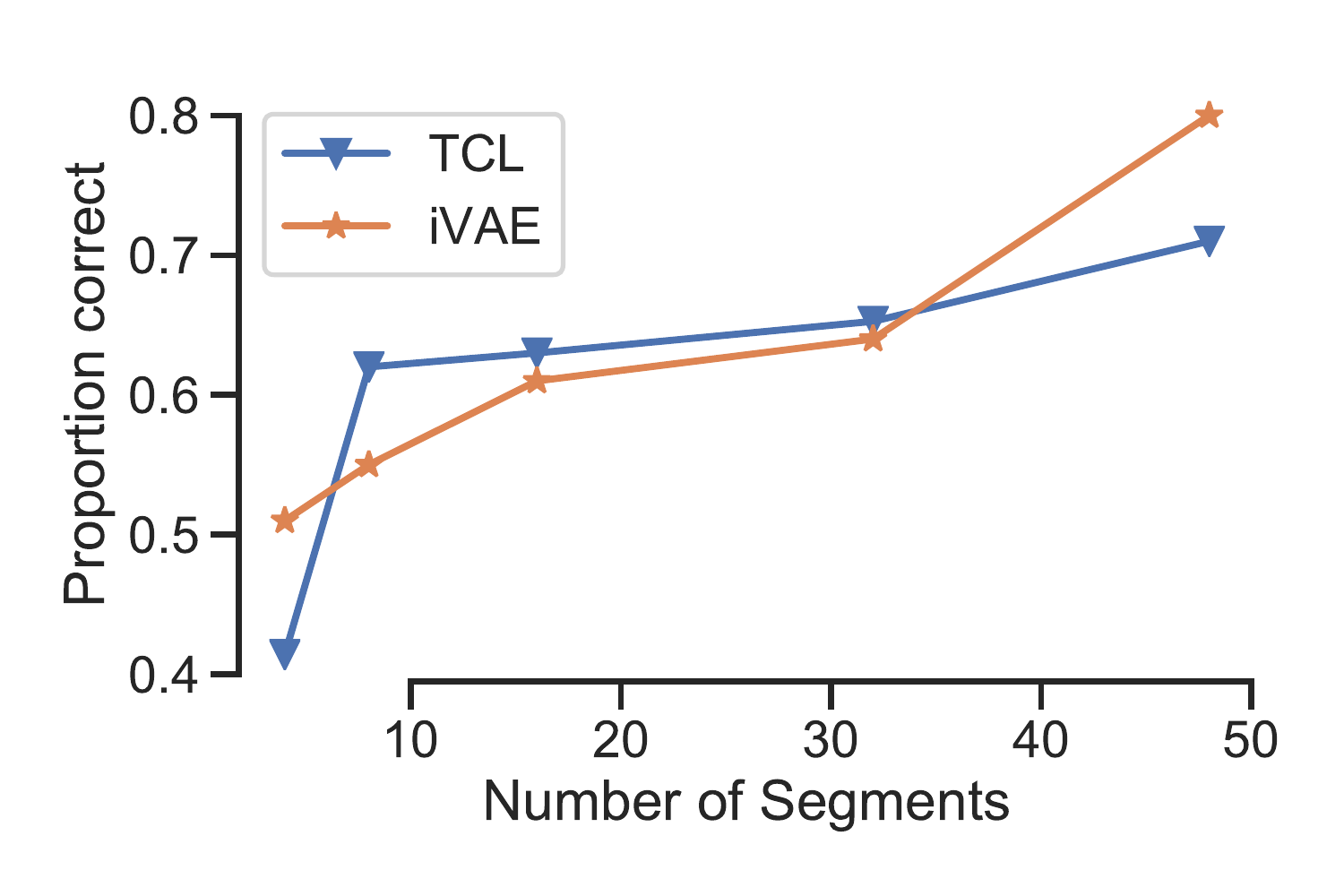}}
    \caption{\label{fig:nonsens_normal} }
\end{subfigure}%
\begin{subfigure}{.32\textwidth}
    \center{\includegraphics[width=\textwidth]
    {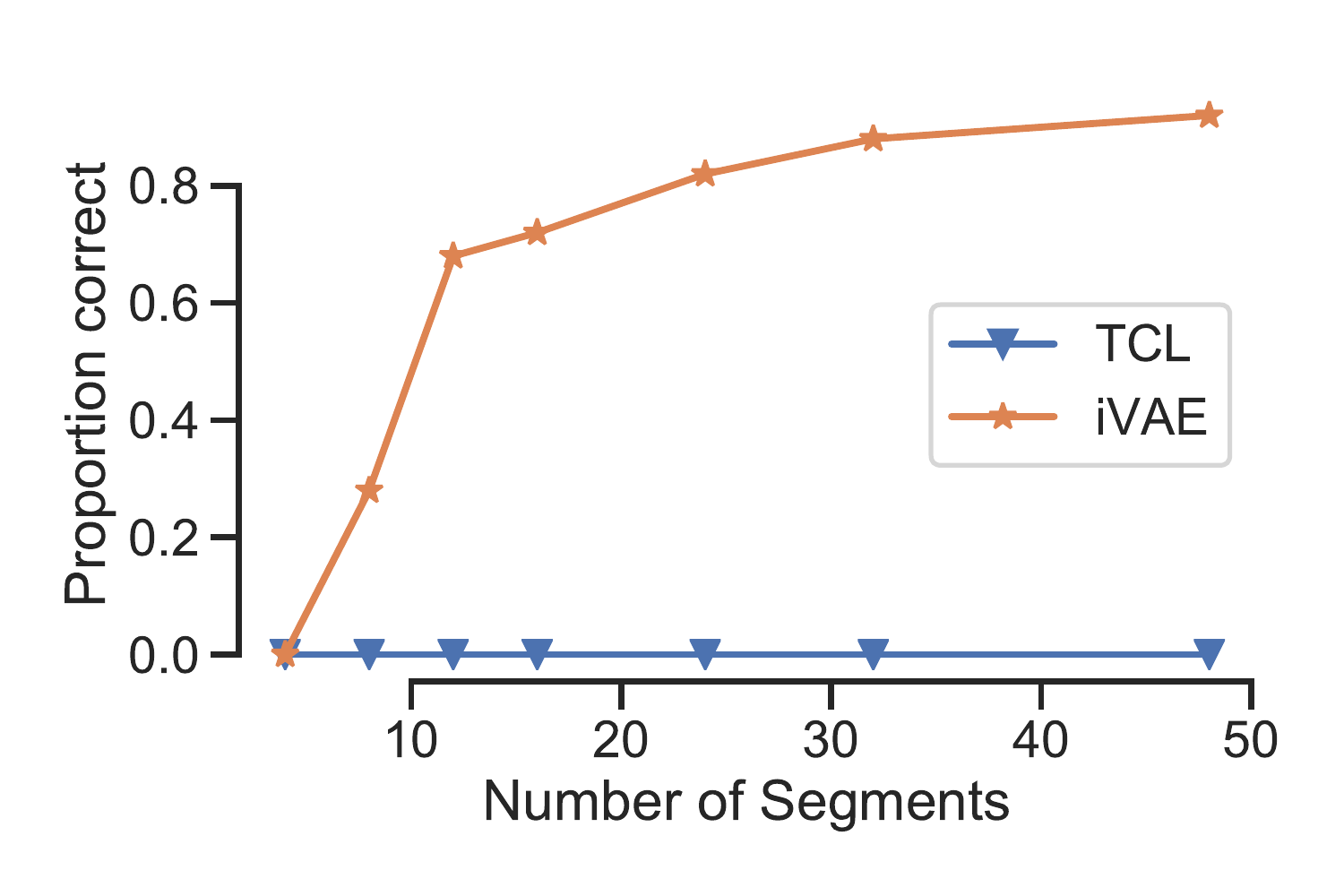}}
    \caption{\label{fig:nonsens_staircase} }
\end{subfigure}%
\begin{subfigure}{.32\textwidth}
    \center{\includegraphics[width=\textwidth]
    {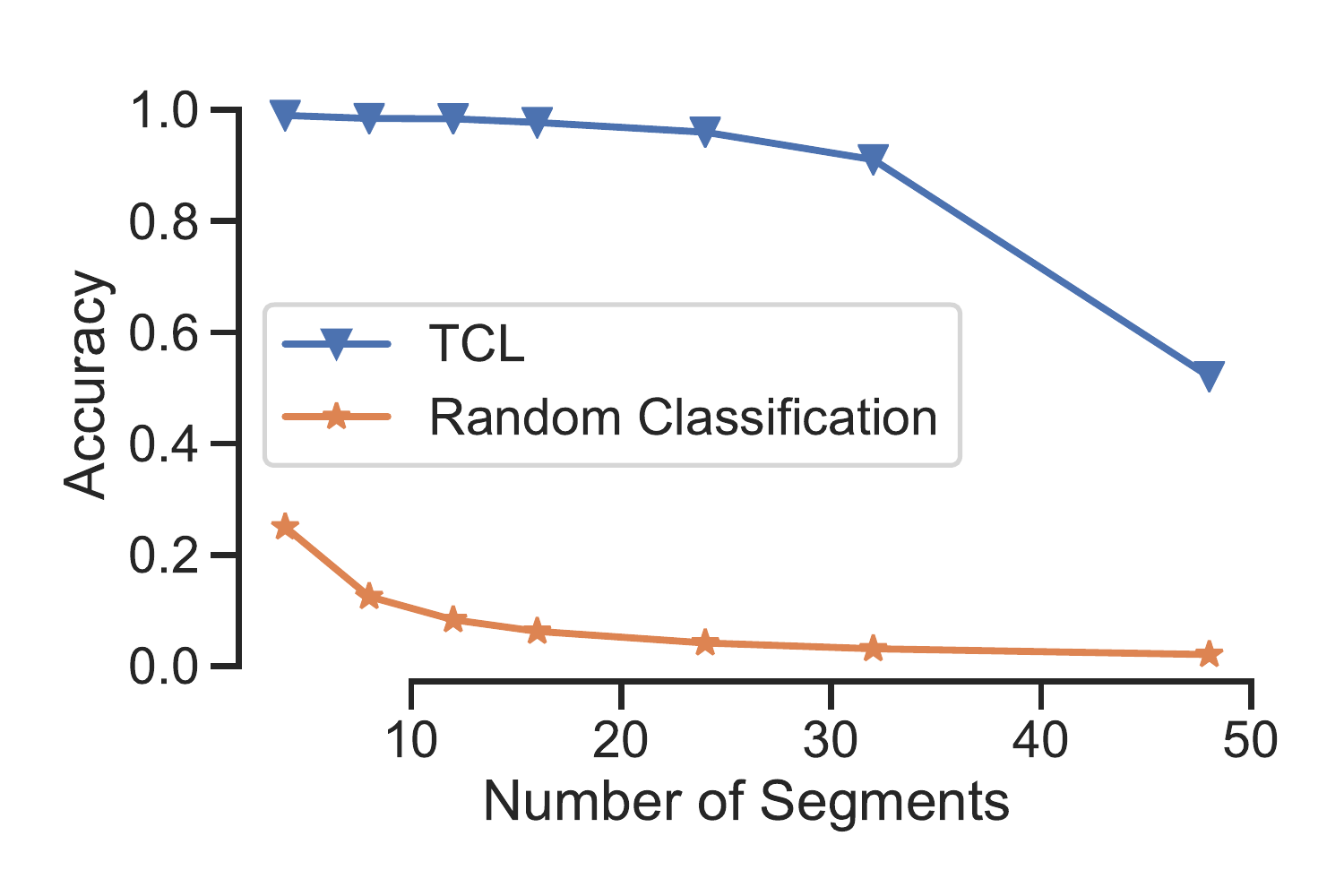}}
    \caption{\label{fig:nonsens_staircase_acc} }
\end{subfigure}
\caption{\label{fig:nonsens} $(a)$ Performance of
nonlinear causal discovery for "normal" data, as described in Section \ref{app:causal}
when iVAE or TCL are employed to recover latent disturbances. $(b)$ Similarly, but when underlying sources 
display significant mean modulation across segments, making them easy to classify. $(c)$ Classification accuracy of TCL when applied on data displaying significant mean modulation. We note that the accuracy of TCL is significantly above a random classifier, indicating that the 
surrogate classification problem employed in TCL training has been effectively optimized.}
\end{figure}

\paragraph{Experiments on "normal" simulated data}
We seek to compare iVAE and TCL in the context of 
causal discovery, as described in 
Section \ref{sec:realdata}. Such an approach involves a two-step 
procedure whereby first 
either TCL or iVAE are employed to recover latent disturbances, followed
by a series of independence tests. 
Throughout all causal discovery experiments we employ
HSIC as a general test of statistical independence 
\citep{gretton2005measuring}.
When comparing iVAE and TCL in this setting we report the 
proportion of times the correct causal direction is reported. It is important to note that the 
aforementioned testing procedure can produce one of three
decisions: $x_1 \rightarrow x_2$, $x_2 \rightarrow x_1$ or 
a third decision which states that no acyclic causal 
direction can be determined. 
The first two outcomes correspond to identifying causal structure and will occur when 
we fail to reject the null hypothesis in only one of the four tests. 
Whereas the third decision (no evidence of acyclic causal structure) will
be reported when either there is evidence to reject the
null in all four tests or we fail to reject the null more than once.
Typically, this will occur if the 
nonlinear unmixing has failed to accurately 
recover the true latent sources. 
The results are reported in Figure \ref{fig:nonsens_normal} 
where we note that both TCL and iVAE perform 
comparably.

\paragraph{Experiments on significant mean modulated data}
As a further experiment, we consider 
causal discovery in the scenario where 
one or both of the underlying sources demonstrate a significant mean 
modulation as shown in Figure \ref{fig:examle_sources}. In such a setting the surrogate classification problem which is solved as part of TCL training becomes significantly easier, to
the extent that TCL no longer needs to learn an accurate 
representation of the log-density of sources within each segment.
This is to the detriment of TCL as it implies that it cannot accurately 
recover latent sources and therefore fails at the task of  
causal discovery. This can be seen in Figure \ref{fig:nonsens_staircase} where TCL based causal discovery fails whereas iVAE continues to 
perform well. This is a result of the fact that iVAE 
directly optimizes the log-likelihood as opposed to a surrogate 
classification problem. 
Moreover, Figure \ref{fig:nonsens_staircase_acc}
visualizes the mean classification accuracy for TCL as a function of the number of segments. We note that TCL consistently obtains 
classification accuracy that are significantly better that random classification. This 
provides evidence that the poor performance of TCL in the context of 
data with significant mean modulations is not a result of sub-optimal optimisation but are instead a a negative consequence of TCL's
reliance on solving a surrogate classification problem to 
perform nonlinear unmixing.

\subsection{Real data experiments}
\label{app:fmri}
\paragraph{Hippocampal fMRI data}
Here we provide further details relating to the resting-state
Hippocampal data provided by 
\cite{poldrack2015longterm} and studied in Section \ref{sec:realdata}, closely following the earlier causal work using TCL by \citet{monti2019causal}.
The data corresponds to daily fMRI scans from a single individual (Caucasian male, aged 45) collected over a period of 
84 successive days. We consider data collected from each day as 
corresponding to a distinct segment, encoded in $\mathbf{u}$.
Within each day 518 BOLD observations are provided across the 
following six brain regions: 
perirhinal cortex (PRc), parahippocampal cortex (PHc), entorhinal cortex (ERc), subiculum
(Sub), CA1 and CA3/Dentate Gyrus (DG).

\subsection{Additional visualisations for comparison to VAE variants}
\label{app:figs}

\begin{figure*}
    \center{\includegraphics[width=.98\textwidth]
    {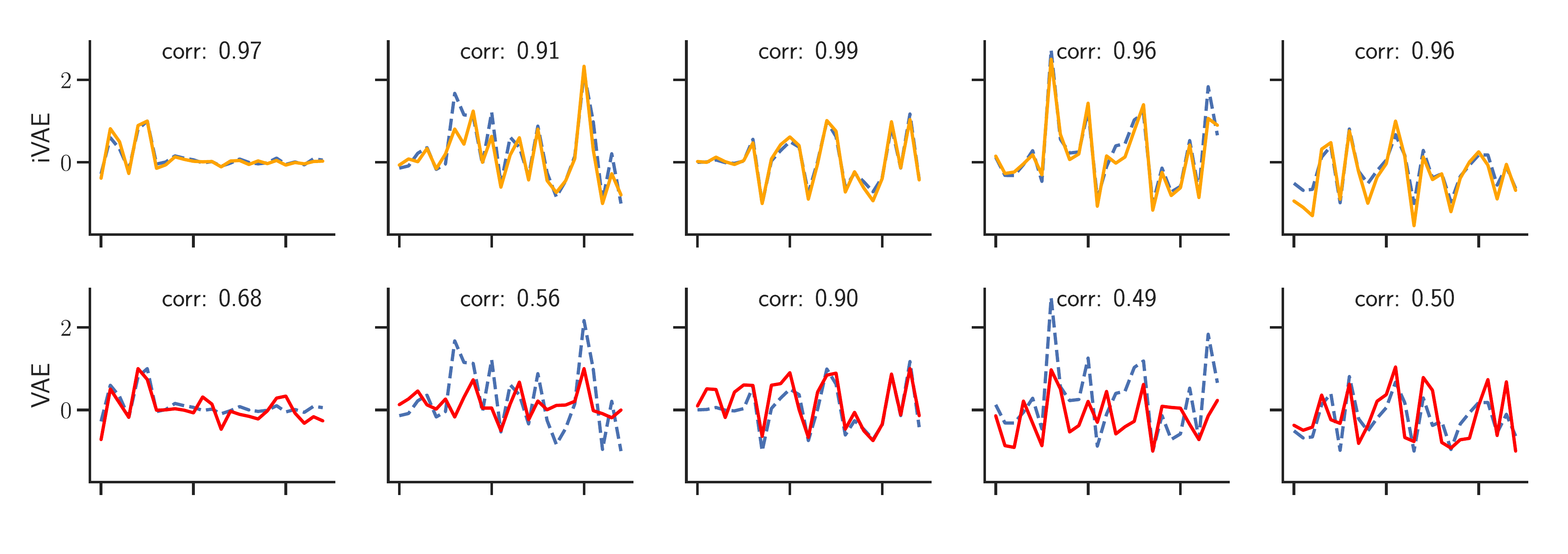}}
    \caption{\label{fig:tcl_recon} Comparison of the recovered latents of our model to the latents recovered by a vanilla VAE. The dashed blue line is the true source signal, and the recovered latents are in solid coloured lines. We also reported the correlation coefficients for every (source, latent) pair. }
\end{figure*}

\begin{figure*}
    \center{\includegraphics[width=.9\textwidth]
    {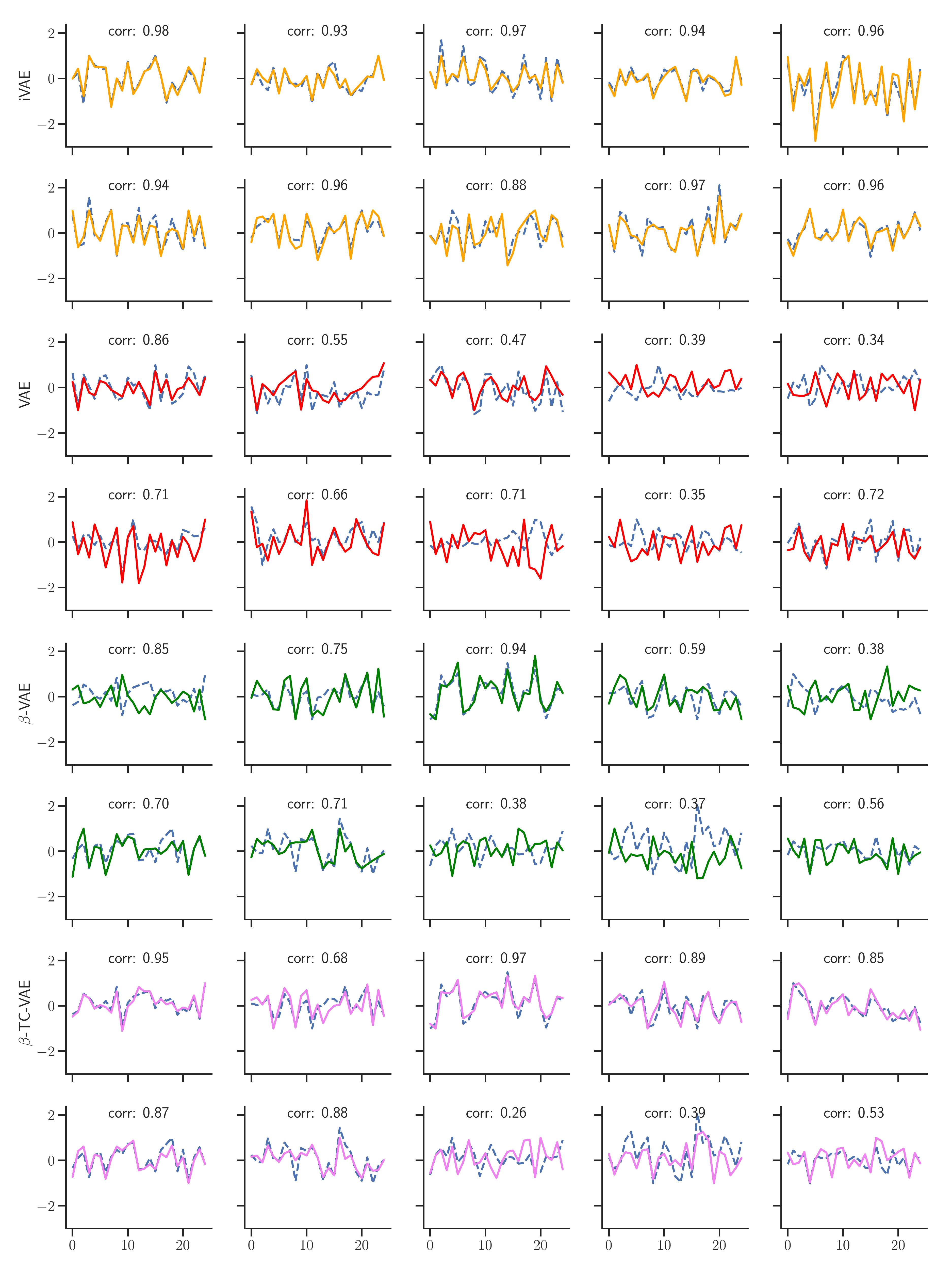}}
    \caption{\label{fig:tcl_recon_d10} Comparison of the recovered latents of our model to the latents recovered by a vanilla VAE, a $\beta$-VAE and a $\beta$-TC-VAE, where the dimension of the data is $d=40$, and the dimension of the latents is $n=10$, the number of segments is $M=40$ and the number of samples per segment is $L=4000$. The dashed blue line is the true source signal, and the recovered latents are in solid coloured lines. We reported the correlation coefficients for every (source, latent) pair. We can see that \VAEICA have an excellent estimation of the original sources compared to the other models.}
\end{figure*}

As a further visualization we show in Figures \ref{fig:tcl_recon} and \ref{fig:tcl_recon_d10} the recovered latents for VAE and \VAEICA; we sampled a random (contiguous) subset of the sources from the dataset, and compared them to the recovered latents (after inverting any permutation in the components). We can see that \VAEICA has an excellent estimation of the original sources compared to  VAE (other models were almost indistinguishable from vanilla VAE).

\section{ACKNOWLEDGEMENT}

I.K.  and R.P.M.  were supported by the Gatsby Charitable Foundation. A.H.  was supported by a Fellowship from CIFAR, and from the DATAIA convergence institute as part of the "Programme d’Investissement d’Avenir", (ANR-17-CONV-0003) operated by Inria.